\documentclass[letterpaper]{article}

\usepackage[T1]{fontenc}
\usepackage{geometry}
\usepackage{setspace}
\usepackage{url}
\usepackage{achemso}            

\usepackage{graphicx}
\usepackage{float}
\usepackage{amsmath}
\usepackage{booktabs}
\usepackage{siunitx}
\usepackage{microtype}
\graphicspath{{figs/}}

\newcommand{\Voc}{V_{\mathrm{OC}}}
\newcommand{\Jsc}{J_{\mathrm{SC}}}
\newcommand{\Ea}{E_{\mathrm{a}}}
\newcommand{\Eg}{E_{\mathrm{g}}}
\providecommand{\texorpdfstring}[2]{#1}
\newcommand{\absorber}{RbSn\texorpdfstring{\textsubscript{0.5}}{0.5}%
Ge\texorpdfstring{\textsubscript{0.5}}{0.5}I\texorpdfstring{\textsubscript{3}}{3}}
\newcommand{\Rtwo}{R^{2}}

\usepackage{authblk}
\author[1]{Nabanita Sarker}
\author[1]{Hafiz Imtiaz*}
\affil[1]{Department of Electrical and Electronic Engineering, Bangladesh University of
Engineering and Technology, Dhaka 1205, Bangladesh}

\title{Interpretable machine learning diagnosis of degradation mechanisms in
lead-free \absorber{} solar cells from temperature- and illumination-dependent electrical
measurements}
\date{*Email: hafizimtiaz@eee.buet.ac.bd}

\begin{document}

\maketitle

\begin{abstract}
Machine learning coupled with device simulation is widely used to optimize thin-film solar cells, but almost exclusively in the forward direction, from device parameters to predicted performance. The inverse problem encountered after fabrication, namely identifying the mechanism that limits a given device, is still addressed by manual inspection of characterization data. In this work, both directions are examined for lead-free \absorber{} solar cells using approximately 42,000 SCAPS-1D simulations. Replacing the Spiro-OMeTAD hole transport layer with CuI increases the simulated power conversion efficiency from 19.6\% to 21.6\%, almost entirely through the fill factor. Surrogate models trained and tested on a factorial parameter grid substantially overstate their predictive accuracy. An XGBoost regressor attains $\Rtwo = 0.9997$ on held-out grid points but only 0.58 on independently sampled devices, whereas training on a Latin hypercube design of comparable size restores $\Rtwo = 0.995$. For the inverse problem, a multi-label classifier identifies which of six degradation mechanisms are present from simulated temperature- and illumination-dependent open-circuit-voltage measurements. It achieves a macro $F_1$ score of 0.938, compared to 0.687 for a calibrated physics-rule baseline. The classifier does not, however, generalize to combinations of mechanisms absent from the training set.
\end{abstract}

\section*{Keywords}
lead-free perovskite; \absorber{}; SCAPS-1D; machine learning; defect diagnosis; ideality factor; activation energy; interpretable models

\section{Introduction}

Lead halide perovskite solar cells now exceed 26\% efficiency \citep{nrel2026chart}. The toxicity of lead, however,
remains an obstacle to their deployment, and regulations on hazardous substances make lead-free
alternatives a practical as well as an environmental requirement. The search for a substitute
has concentrated on the group-14 congeners of lead, and on tin in particular. The reason is that
Sn\textsuperscript{2+} reproduces the electronic structure that makes the lead compounds
effective absorbers. \citet{stoumpos2013} characterized the tin and lead iodide perovskites side
by side. They reported high carrier mobilities and narrower band gaps for the tin compounds,
which places these compounds closer to the single-junction optimum. Reviews of the field
nevertheless record a persistent gap between this promise and the measured device performance
\citep{saranya2023}. For example, a CsSnI\textsubscript{3} cell without a hole transport layer
reached 3.56\%, the highest efficiency reported for that absorber at the time
\citep{marshall2016}. The more air-stable, vacancy-ordered
Cs\textsubscript{2}SnI\textsubscript{6} has been proposed precisely because the parent compound
degrades so rapidly \citep{qiu2017}. The cause of this gap is chemical rather than
optoelectronic. Sn\textsuperscript{2+} readily oxidizes to Sn\textsuperscript{4+}, which
self-dopes the absorber and introduces deep traps. Tin-based devices therefore lose performance
on a timescale over which lead-based devices remain stable.

Alloying tin with germanium addresses the oxidation pathway directly rather than merely slowing
it. \citet{ju2017} predicted that germanium in a mixed Sn--Ge sublattice oxidizes preferentially
and forms a thin native oxide that protects the underlying material. \citet{chen2019}
demonstrated this effect experimentally in all-inorganic lead-free cells and attributed their
stability to this native-oxide passivation. The tin--germanium family has since been reviewed as
a distinct route to stable lead-free absorbers \citep{azizman2023}. Mixed-cation compositions
have also been explored to balance the band gap against stability \citep{azizman2024}. These
approaches complement the surface and bulk passivation strategies developed separately for lead
and tin perovskites. Such strategies include ligand-assisted control of crystallization
\citep{zhu2022} and SnX\textsubscript{2} additives that modulate the defect population in
FASnI\textsubscript{3} \citep{joy2024}. They also include surface treatments that stabilize
inorganic CsPbI\textsubscript{3} \citep{li2018} and simultaneous passivation of the bulk and the
surface \citep{mahmud2022}. None of these strategies, however, removes the need for an absorber that is
intrinsically less prone to oxidation.

\absorber{} lies at the favorable end of this family. It combines germanium stabilization with
a direct band gap of approximately \SI{1.2}{\electronvolt}, strong absorption and low carrier
effective masses. It has therefore been identified computationally as a promising absorber
\citep{gomathi2023,laouid2024}. Related rubidium and cesium germanium compositions have been
examined in the same way. \citet{rbgei3_2022} optimized an inorganic cell based on
RbGeI\textsubscript{3}, the germanium-only end member, to an efficiency of 10.11\%.
\citet{nazari2025} combined Cs\textsubscript{0.5}Sn\textsubscript{0.5}GeI\textsubscript{3} with
an alloyed ZnO\textsubscript{0.3}S\textsubscript{0.7} electron transport layer. No experimental
\absorber{} device has yet been reported. Every quantitative statement about this absorber,
including those made in this work, therefore rests on simulation.

This situation is typical for a candidate absorber, and device simulation with SCAPS-1D
\citep{burgelman2000} is the standard method for assessing one. \citet{yasin2021} designed a
lead-free heterostructure and raised the efficiency of their simulated cell from 14.34\% to
24.81\%. They achieved this by lowering the absorber defect density to
\SI{1e14}{\per\centi\metre\cubed} and tuning the layer thicknesses. \citet{cspbi3_cssni3_2026}
reported 31.45\% for a CsPbI\textsubscript{3}/CsSnI\textsubscript{3} dual-absorber stack. The
same approach is used to screen individual transport materials and to carry first-principles
predictions through to a device-level estimate \citep{teimouri2018,rafi2026}. Screening of
device architectures is equally common. For example, \citet{vaishnavi2025} compared twenty-one
MASnI\textsubscript{3} configurations using three electron transport and six hole transport
materials, with efficiencies ranging from 14.28\% to 24.28\%. The method is not purely
speculative. \citet{yadav2022} fitted a simulated SnSe device to their own measured $J$--$V$ and
quantum-efficiency curves before extrapolating to 22.69\%. However, \citet{saidarsan2025}
surveyed approximately 250 published SCAPS studies. They found unrealistic parameter choices and
results to be common enough to warrant a corrective review. For this reason, the baseline used
here is checked against its published source before any new result is generated.

Because each SCAPS evaluation carries a small but non-negligible computational cost, machine
learning is increasingly used to accelerate the associated parameter searches.
\citet{mustafa2025} simulated 56,390 CsPbI\textsubscript{3} configurations and trained five
regressors, of which XGBoost reached $\Rtwo = 0.9998$ on held-out test data. They used SHAP to identify absorber
thickness and defect density as the key parameters, and they raised the predicted efficiency
from 15.15\% to 19.16\%. \citet{ikram2026} applied the same framework to
Sb\textsubscript{2}S\textsubscript{3}. They improved a 9.86\% baseline to 17.69\% and identified
a critical bulk defect density of \SI{1e16}{\per\centi\metre\cubed}. \citet{malek2026}
optimized efficiency and stability together in a cell without a hole transport layer. They
raised the efficiency from 13.7\% to 16.84\% and reduced the degradation over \SI{1000}{\hour}
from 6.61\% to 2.39\%. \citet{sulfide_ml2025} improved a sulfide reference cell from 22.96\% to
31.43\% by the same approach. \citet{jiang2020} used machine learning to guide the KI doping of
MAPbI\textsubscript{3} and fabricated a best device with an efficiency of 20.91\%. Finally,
\citet{odabasi2019} applied the approach to published rather than simulated data and mined the
literature for the factors that govern perovskite stability.

Gradient-boosted trees \citep{chen2016} are the usual choice of model. They remain competitive
with neural networks when only a small number of simulations is available for training, as
comparisons in other simulation-driven domains also show \citep{stevanovic2024}. The workflow is
uniform across this literature. A grid of device parameters is simulated, a regressor is trained
to predict the photovoltaic outputs, and SHAP \citep{lundberg2017} or a similar method is used to
rank the inputs. Two consequences of this uniformity motivate the present work.

The first consequence concerns the evaluation of such surrogates. When the training and test
devices are drawn from the same parameter grid, the held-out accuracy measures agreement at the
grid nodes. It does not measure prediction at parameter combinations that the grid never
visited. Tree ensembles are, moreover, known to extrapolate poorly beyond the region on which
they were fitted \citep{numata2023}. Space-filling designs \citep{mckay1979} avoid the
axis-aligned structure that creates this problem. Nevertheless, the grid protocol remains the
default, and the accuracy it reports is the value that is subsequently quoted.

The second consequence concerns a question that the forward workflow does not address.
Predicting the outputs of a parameter set answers the question of \emph{which parameters a
fabricated device should aim for}. It does not answer the question that an experimentalist faces
after fabrication, namely \emph{which mechanism is degrading the device at hand}. This inverse
question has a long-established experimental basis \citep{hegedus2004,tress2018}. The
temperature dependence of $\Voc$ yields an activation energy, and a deficit of this activation
energy relative to the band gap indicates interface rather than bulk recombination. The ideality
factor, obtained from the light-intensity dependence of $\Voc$, distinguishes trap-assisted from
band-to-band recombination. Both measurements are in routine experimental use.
\citet{kim2017} attributed the large $\Voc$ loss of
CH\textsubscript{3}NH\textsubscript{3}SnI\textsubscript{3} cells to interface rather than bulk
recombination on exactly this evidence. \citet{stolterfoht2020} used related measurements to
separate the efficiency potential of a perovskite film from the losses imposed by its contacts.
\citet{ebadi2025} combined temperature-dependent $\Voc$, quantum efficiency and luminescence to
identify tail states in a double perovskite. These signatures are nevertheless interpreted by
inspection, one device at a time \citep{kumar2021}. They also become ambiguous when several loss
mechanisms act together, which is the usual condition in a fabricated cell.

This work addresses both gaps. We first show that the standard evaluation protocol for
grid-trained surrogates substantially overstates their accuracy. We also show that space-filling
sampling corrects this problem at a lower computational cost. We then train models to invert the
problem. From simulated measurements that are experimentally accessible, these models identify
which of six modeled degradation mechanisms are present in a device and estimate the severity of the
degradation. Together with the healthy reference, this gives seven diagnostic classes. The
models are benchmarked against a calibrated implementation of the established physics rules.
They are tested against measurement noise and outside their training range. Finally, they are
applied to published measurements of tin perovskite devices fabricated by other groups.

The specific contributions of this work are as follows.
\begin{enumerate}
    \item \textbf{The sampling-design problem.} A quantification of this problem in SCAPS-based surrogate modeling: the same model, trained on the same grid devices, attains $\Rtwo = 0.9997$ on grid test points but only 0.579 on independently sampled devices.

    \item \textbf{An interpretable diagnostic framework.} The framework identifies the degradation mechanisms of a device from $\Voc(T)$ and $\Voc(\Phi)$, achieving a macro $F_1$ score of 0.938 compared with 0.687 for calibrated physics rules. On single-degradation devices, its forced-choice accuracy is 99\% compared with 61\%. On devices with two simultaneous mechanisms, its exact-match accuracy is 69\% compared with 22\%.

    \item \textbf{Practical requirements for applying such models to measured data.} Training must include measurement noise, and the analyzed measurement range must match the training range. Abstention should be based on model confidence rather than on the quality of the extracted fits.

    \item \textbf{A loss-channel attribution of the CuI gain.} This attribution concerns the efficiency gain obtained by replacing Spiro-OMeTAD with an inorganic CuI hole transport layer in a TiO\textsubscript{2}-based \absorber{} stack, and separates the contributions of series resistance, absorber defects and hole mobility. \citet{amin2026} independently identified CuI as the preferred hole transport layer for this absorber by screening five candidates. The decomposition reported here complements that screening and is not a claim of priority.
\end{enumerate}
\section{Methods}

\subsection{Device structure and simulation settings}
\label{sec:methods_device}

\begin{figure}[tbp]
\centering
\includegraphics[width=0.80\textwidth]{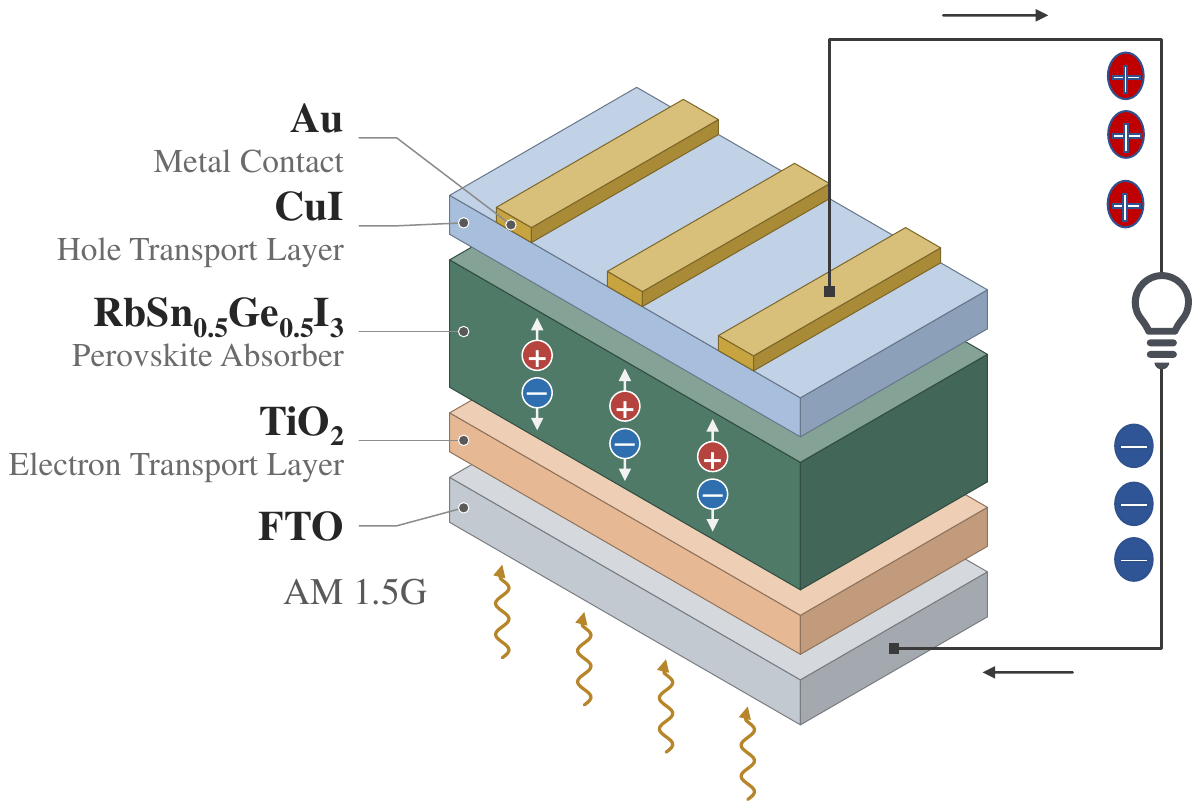}
\caption{Schematic of the simulated FTO/\allowbreak TiO\textsubscript{2}/\allowbreak\absorber{}/\allowbreak CuI/\allowbreak Au device. Light enters
through the FTO front contact; electrons are collected at the FTO contact and holes at the Au
contact. Layer thicknesses are not to scale. In the simulation, the p-side is placed at the left
of the SCAPS domain, as recommended in the SCAPS manual for numerical stability, and the Au
contact is represented as a boundary condition rather than as a layer of finite thickness.}
\label{fig:schematic}
\end{figure}

A planar n--i--p device with the structure FTO/\allowbreak TiO\textsubscript{2}/\allowbreak\absorber{}/\allowbreak CuI/\allowbreak Au was modeled with SCAPS-1D (version 3.3.12) \citep{burgelman2000}, with TiO\textsubscript{2} as the electron transport layer (ETL) and CuI as the hole transport layer (HTL). Unless stated otherwise, all simulations were performed under AM1.5G illumination at \SI{100}{\milli\watt\per\centi\metre\squared} and at a temperature of \SI{300}{\kelvin}. Light enters through the FTO contact, and the p-side of the structure is placed at the left of the simulation domain (Fig.~\ref{fig:schematic}), as recommended in the SCAPS manual for numerical stability.

The absorber, TiO\textsubscript{2} and FTO parameters were taken from \citet{gomathi2023}, who studied the same absorber with a Spiro-OMeTAD HTL, and the CuI parameters follow \citet{amin2026}. Because the relative permittivity of the absorber is not reported in the 2023 study, the value of 6.5 was taken from \citet{laouid2024, amin2026}. The complete parameter set is listed in Table~\ref{tab:params}.

Recombination in the absorber comprises Shockley--Read--Hall (SRH), radiative
(\SI{1e-11}{\centi\metre\cubed\per\second}) and Auger (\SI{1e-28}{\centi\metre\tothe{6}\per\second}) channels. Bulk defects are neutral, single-level and located at midgap ($E_t = \SI{0.6}{\electronvolt}$ above $E_V$), with capture cross-sections $\sigma_n = \sigma_p = \SI{1e-15}{\centi\metre\squared}$. Each heterointerface contains a neutral midgap defect with the same cross-sections, specified \emph{on both sides of the interface}. This specification is required because SCAPS assigns separate cross-sections to carriers arriving from the left and right layers, and leaving either side at its default value
suppresses recombination at that interface. The baseline series and shunt resistances are 2 and \SI{1000}{\ohm\centi\metre\squared}, respectively, and the baseline Au work function is \SI{5.0}{\electronvolt}; all three quantities are varied from device to device in the diagnostic dataset of Section~\ref{sec:datasets}. Because the absorption data used in the reference study are not tabulated in the source publication, the absorption coefficient follows the SCAPS $\sqrt{h\nu - \Eg}$ model with $\alpha_0 = \SI{1e5}{\per\centi\metre}$.

\citet{saidarsan2025} surveyed approximately 250 published SCAPS-1D perovskite architectures and found that a substantial fraction report efficiencies exceeding the Shockley--Queisser limit for the stated band gap, an outcome they attributed mainly to the radiative recombination coefficient being left at its SCAPS default of zero. They accordingly recommend that this coefficient be stated explicitly and that simulated results be checked against the limit. In
the present work, the radiative coefficient is \SI{1e-11}{\centi\metre\cubed\per\second} for both the absorber and CuI (Table~\ref{tab:params}), and the detailed-balance limit was evaluated from
the same AM1.5G spectrum file that SCAPS reads. At $\Eg = \SI{1.2}{\electronvolt}$, the limit corresponds to $\Voc = \SI{0.950}{\volt}$,
$\Jsc = \SI{38.71}{\milli\ampere\per\centi\metre\squared}$, $\mathrm{FF} = 87.7\%$ and $\eta = 32.3\%$; as a check, the same calculation yields 33.5\% at \SI{1.34}{\electronvolt}, compared with the commonly quoted value of 33.7\%. No device in any of the datasets generated here exceeds the limit in any of these four quantities. The highest values encountered are $\Voc = \SI{0.909}{\volt}$, $\Jsc = \SI{37.91}{\milli\ampere\per\centi\metre\squared}$,
$\mathrm{FF} = 84.1\%$ and $\eta = 27.3\%$, and the best SCAPS-verified candidate of Section~\ref{sec:optimisation} reaches 71\% of the detailed-balance efficiency.

\begin{table}[tbp]
\centering
\caption{Baseline layer parameters of the simulated FTO/\allowbreak TiO\textsubscript{2}/\allowbreak\absorber{}/\allowbreak CuI/\allowbreak Au device,
read from the SCAPS definition file used for all simulations. A value of zero indicates that the mechanism is disabled for that layer, not a missing value. Several parameters are overridden in individual runs, as described in Section~\ref{sec:datasets}. Absorber, TiO\textsubscript{2} and FTO values follow \citet{gomathi2023}, and CuI values follow \citet{amin2026}; the absorber permittivity, which \citet{gomathi2023} do not report, is taken from \citet{laouid2024, amin2026}.}
\label{tab:params}
\begin{tabular}{@{}lrrrr@{}}
\toprule
Parameter & FTO & TiO\textsubscript{2} & \absorber{} & CuI \\
\midrule
Thickness (\si{\micro\metre})                       & 0.4 & 0.02 & 0.35 & 0.1 \\
Band gap $\Eg$ (\si{\electronvolt})                  & 3.3 & 3.0 & 1.2 & 3.1 \\
Electron affinity $\chi$ (\si{\electronvolt})       & 4.1 & 4.0 & 4.0 & 2.1 \\
Relative permittivity $\varepsilon_r$               & 9 & 9 & 6.5 & 6.5 \\
$N_C$ (\si{\per\centi\metre\cubed})                 & \num{1.2e18} & \num{1.2e19} & \num{2e18} & \num{2.8e19} \\
$N_V$ (\si{\per\centi\metre\cubed})                 & \num{1.8e19} & \num{1.8e19} & \num{1e19} & \num{1e19} \\
$\mu_e$ (\si{\centi\metre\squared\per\volt\per\second}) & 50 & 0.001 & 20 & 100 \\
$\mu_h$ (\si{\centi\metre\squared\per\volt\per\second}) & 50 & 0.001 & 20 & 43.9 \\
$N_A$ (\si{\per\centi\metre\cubed})                 & 0 & 0 & \num{1e15} & \num{1e18} \\
$N_D$ (\si{\per\centi\metre\cubed})                 & \num{1e18} & \num{1e18} & 0 & 0 \\
Radiative coeff. (\si{\centi\metre\cubed\per\second}) & 0 & 0 & \num{1e-11} & \num{1e-11} \\
Auger coeff. (\si{\centi\metre\tothe{6}\per\second})  & 0 & 0 & \num{1e-28} & 0 \\
\bottomrule
\end{tabular}
\end{table}

\subsection{Baseline check against the reference simulation}

Before any data were generated, the Spiro-OMeTAD reference device of \citet{gomathi2023} was compared with the values reported in that study: an optimum absorber thickness near \SI{300}{\nano\metre}, with $\Voc = \SI{0.855}{\volt}$, a $\Jsc$ saturating at \SI{38.5}{\milli\ampere\per\centi\metre\squared}, an FF of 82.36\% and an efficiency of 27.1\%. For a bulk defect density of \SI{1e14}{\per\centi\metre\cubed}, our model of the same structure reaches an efficiency of 20.2\% at \SI{0.6}{\micro\metre}. At the reported optimum of \SI{300}{\nano\metre}, it yields $\Voc = \SI{0.887}{\volt}$, $\Jsc = \SI{32.0}{\milli\ampere\per\centi\metre\squared}$, an FF of 67.2\% and an efficiency of
19.1\%. This comparison is made against the reported values and does not constitute a reproduction of the published figures: Fig.~4 of \citet{gomathi2023} is a one-dimensional thickness sweep and its Fig.~5 a contour plot against carrier \emph{lifetime}, whereas the second
axis in the present analysis is the bulk defect density, which is related to, but not identical with, the lifetime.

The fill-factor difference was apportioned among individual loss channels by removing each channel separately from the same baseline. Series resistance accounts for 6.0 points, absorber defects for 5.2 points, the low hole mobility of Spiro-OMeTAD (\SI{1e-4}{\centi\metre\squared\per\volt\per\second}) for 7.2 points and shunt leakage for 1.0 point, whereas the TiO\textsubscript{2} mobility contributes 0.05 points, which is not a measurable effect. Removing the channels cumulatively yields a total of 20.7 points, compared with 19.4 points for the sum of the individual contributions. The channels therefore interact,
and neither figure constitutes an exact decomposition. Series and shunt resistance are reported separately because their contributions differ by a factor of six.

With every tested loss channel removed, the model reaches 26.0\%, compared with the reported 27.1\%. This agreement, however, arises from compensating discrepancies: our $\Jsc$ is \SI{6.4}{\milli\ampere\per\centi\metre\squared} lower than the reported value, whereas our FF is 5.6 points higher and our $\Voc$ is \SI{67}{\milli\volt} higher. The $\Jsc$ shortfall at \SI{300}{\nano\metre} is the largest single difference and is unaffected by any of the tested loss channels. Our $\Jsc$ nonetheless reaches \SI{37.9}{\milli\ampere\per\centi\metre\squared} at \SI{1.0}{\micro\metre}, close to the reported saturation value, which points to the absorption model, rather than to recombination or transport, as the source of the difference. Table~1 of \citet{gomathi2023} attributes the absorption of the absorber to \citet{ju2017} without specifying a functional form, and its text describes the absorption as constant and uniform, with an internal quantum efficiency of 100\%. The present model, in contrast, uses the SCAPS $\sqrt{h\nu - \Eg}$ law with $\alpha_0 = \SI{1e5}{\per\centi\metre}$. This difference cannot be
tested through the script interface, which does not expose the absorption parameters. The model therefore reproduces the qualitative thickness dependence of the reference device and its $\Jsc$ at large thickness, but not the published photovoltaic parameters quantitatively, and the remaining difference is not attributed to a single cause.

\begin{figure}[tbp]
\centering
\includegraphics[width=\textwidth]{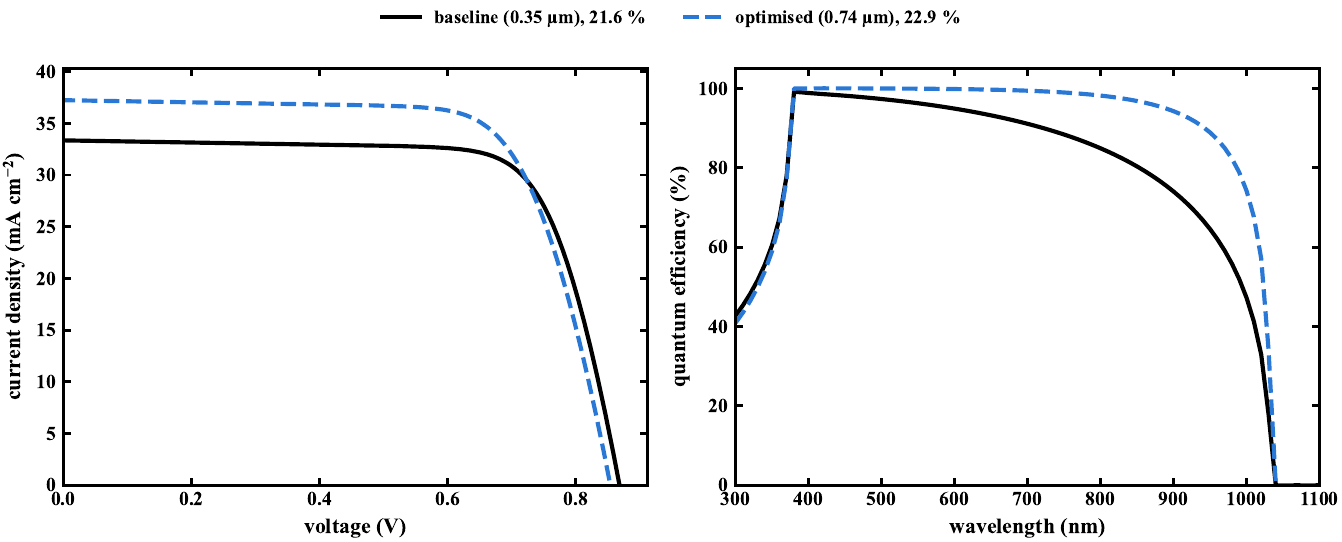}
\includegraphics[width=0.49\textwidth]{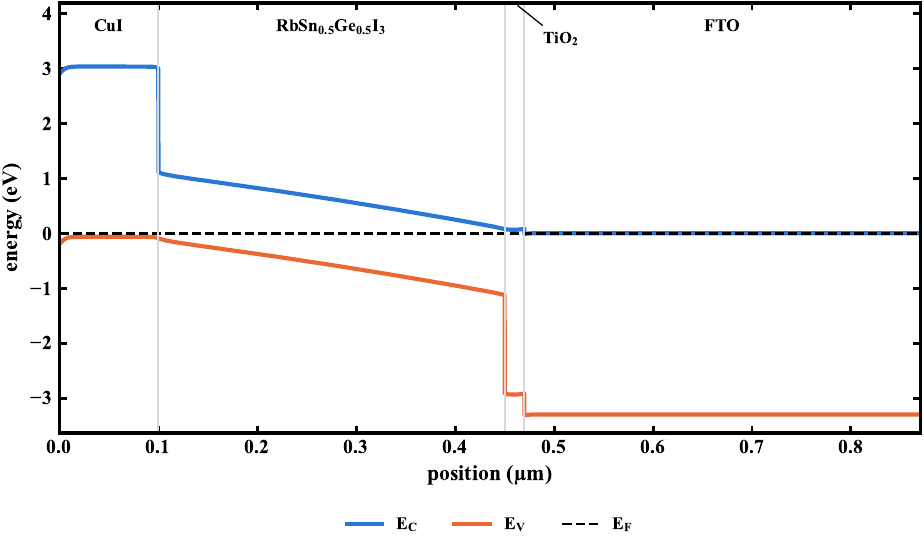}
\includegraphics[width=0.49\textwidth]{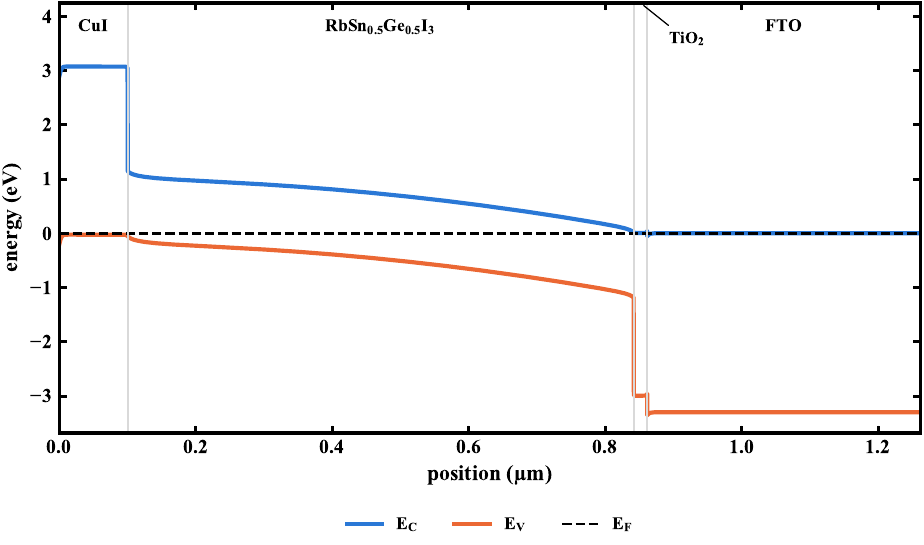}
\caption{Baseline FTO/\allowbreak TiO\textsubscript{2}/\allowbreak\absorber{}/\allowbreak CuI/\allowbreak Au device, including the interface defects described in Section~\ref{sec:methods_device}. Top: simulated $J$--$V$ curves and external quantum efficiency of the baseline device (21.6\%) and the surrogate-optimized device (22.9\%). Bottom: equilibrium band diagrams of the baseline device (left) and the surrogate-optimized device (right).}
\label{fig:device}
\end{figure}

Replacing Spiro-OMeTAD with CuI removes most of the transport loss: at an absorber thickness of \SI{0.35}{\micro\metre}, the fill factor increases from 68.1\% to 74.6\% and the efficiency from 19.6\% to 21.6\%. Both HTL configurations are defined through the same set of layer parameters,
so that the two devices are matched in every property controlled by the script. The comparison is made on the final device definition, which includes the interface defects described in Section~\ref{sec:methods_device}; the CuI device in this comparison is therefore the baseline for every dataset in Section~\ref{sec:datasets} and the device shown in Fig.~\ref{fig:device}.
Reported values of the absorber mobility range from \SIrange{2}{20}{\centi\metre\squared\per\volt\per\second}. Switching between the two values changed
the efficiency of either device by less than 0.05 percentage points, so the results are insensitive to this choice. The resulting $J$--$V$ and quantum-efficiency curves are shown in Fig.~\ref{fig:device}, together with the equilibrium band diagrams.

\subsection{Dataset generation}\label{sec:datasets}
All simulations were performed through the SCAPS script interface, with several instances running in parallel; a single $J$--$V$ calculation requires approximately \SI{1.5}{\second}. In total, 42,238 simulations were attempted, of which 297 (0.7\%) failed to converge, all within the diagnostic datasets described below. Every value reported in this work is obtained from a
converged run, and a per-batch ledger that reconciles this total with the stored run records is included with the data (see Data availability).

\paragraph{Forward dataset (Part 1)} Six inputs were varied: the absorber thickness (\SIrange{0.2}{1.0}{\micro\metre}), the absorber bulk defect density
(\SIrange{1e14}{1e17}{\per\centi\metre\cubed}), the interface defect densities at both heterojunctions (\SIrange{1e10}{1e13}{\per\centi\metre\squared}), and the doping densities of TiO\textsubscript{2} and CuI (\SIrange{1e16}{1e19}{\per\centi\metre\cubed}). Three datasets were generated: a full factorial grid with four levels per input (4,096 devices), a space-filling Latin hypercube set of 3,000 devices \citep{mckay1979}, and an independent Latin hypercube test
set of 300 devices drawn with a different random seed. All of these simulations converged. In both the grid and the space-filling datasets, most of the linear signal is carried by the bulk defect density and the absorber thickness, whereas the transport-layer doping densities are nearly uncorrelated with the outputs, which accounts for the poor performance of the linear model in Table~\ref{tab:matched}.

\paragraph{Diagnostic dataset (Part 2)} Each device was assigned labels from seven diagnostic classes: a healthy reference class and six introduced degradation mechanisms, namely bulk defects, ETL-side interface defects, HTL-side interface defects, a back-contact barrier, low absorber mobility and parasitic resistance. A mechanism was introduced by degrading the
corresponding parameter, while all remaining parameters were drawn at random from their healthy ranges. A label therefore records the parameter that was deliberately degraded, which is not necessarily the mechanism that dominates the losses of that device.

This correspondence was verified directly by simulation. For 72 single-degradation training
devices (twelve per class), only the labeled parameter was restored to the middle of its healthy
range. Depending on the class, this restoration raises the efficiency by 1.8--11.8 percentage
points and brings 97\% of the devices within the efficiency range of the healthy population and
92\% above its 5th percentile. In 8\% of cases, the gain is below one percentage point, indicating
that the labeled parameter was degraded but has little effect on the efficiency. The restoration
is a defined intervention rather than a reconstruction of the original healthy value of the
device, which the generator overwrites. It therefore quantifies the efficiency cost of the
intervention and does not attribute the total losses of a device to a single mechanism.

The labels are a weaker indicator of the dominant loss when several mechanisms coexist. In 39
multi-defect devices, each label was restored individually, and the result was compared with
restoring all labels together. A single label accounts for a median of 0.40 of the recoverable
efficiency loss, and 37\% of the labels account for less than one quarter of it. The two
interface classes are the weakest, with median fractions of 0.13 for the ETL side and 0.005 for
the HTL side, compared with 0.71 for bulk defects and 0.64 for the back barrier. The HTL
interface also produces the smallest efficiency loss when it acts alone (mean gain 1.8 points,
negligible in one third of the devices), which is consistent with its being the class that the
diagnostic model identifies least reliably (Section~\ref{sec:diagnosis}).

Two classes represent grouped interventions rather than single physical parameters. The
parasitic-resistance class degrades the series and shunt resistances together, so the framework
identifies resistive losses as a group and cannot attribute a device to series or shunt
resistance individually. This grouping entails a genuine loss of information, since removing the
two resistances independently from the reference device changes the fill factor by 6.0 and 1.0
points, respectively. Similarly, the low-mobility class sets the electron and hole mobilities
together, so that restoring this class tests a combined transport intervention rather than the
transport of a single carrier type.

The background parameters carry no label information on their own: a classifier trained on these
parameters alone reaches a macro-averaged $F_1$ score (macro $F_1$) of only 0.031. Thirty percent
of the devices received two simultaneous defects, which makes the task a multi-label
classification problem. Each device was simulated at five temperatures
(\SIrange{300}{400}{\kelvin} at 1 sun) and at five light intensities (0.01--1 sun at
\SI{300}{\kelvin}); because the standard condition is shared, this amounts to nine runs per
device. Temperatures below approximately \SI{290}{\kelvin} were avoided because SCAPS convergence
deteriorates in that range, as also noted in the reference study.

A second, more demanding dataset was generated after the initial analysis showed that measured
devices are considerably more degraded than the simulated population. In this dataset, bulk
defect densities extend to \SI{1e18}{\per\centi\metre\cubed}, interface defect densities to
\SI{1e14}{\per\centi\metre\squared} and back-contact work functions down to
\SI{4.0}{\electronvolt}, and 55\% of the devices exhibit two or three simultaneous losses. The two
diagnostic datasets comprise 2,000 and 1,200 training devices and 300 and 200 test devices,
respectively. In the first dataset, 0.15\% of the runs failed to converge, compared with 2.2\% in
the harder dataset, mostly at 0.01 sun. Devices with any failed condition were discarded, which
removed 6\% of the intended population, so the coverage of the most degraded devices is limited
by convergence rather than by design.
\subsection{Features and models}
\label{sec:features}

The forward and inverse directions share the same device structure and simulation settings but
use separate datasets. The forward direction follows the established workflow, in which device
parameters are simulated and a surrogate model predicts the photovoltaic outputs. The inverse
direction starts from curves that can be measured in the laboratory, reduces them to physically
interpretable descriptors, and returns the degradation mechanisms present together with their
severity.

The forward models predict $\Voc$, $\Jsc$, FF and PCE from the six device parameters, with the
defect and doping densities log-transformed. Five regressors were compared: linear regression,
quadratic regression, random forest, XGBoost \citep{chen2016} and a multilayer perceptron (MLP).
The models were evaluated with a 75/25 train--test split and with repeated five-fold
cross-validation, and they were interpreted with SHAP \citep{lundberg2017}. XGBoost was used
through its own library, and all other models were implemented with scikit-learn
\citep{pedregosa2011}.

The diagnostic models use only quantities that correspond to laboratory measurements. Every input
is derived from simulated $J$--$V$ curves at a set temperature and illumination intensity, as
obtainable with a solar simulator equipped with a temperature stage and neutral-density filters,
and no internal simulation variable, such as the carrier density, the recombination rate or the
band profile, is used. The inputs are nevertheless not direct readings: the activation energy,
the ideality factor and the various slopes are \emph{inferred} from measured points through a
fitting model and inherit its assumptions. In particular, $\Ea$ is the zero-temperature intercept
of a linear fit to $\Voc(T)$ between 300 and \SI{400}{\kelvin}. This intercept equals the
activation energy of the dominant recombination path only if that path remains dominant across
the fitted range and the temperature dependence of the band gap and of the effective densities of
states is negligible. The quality of the fit is therefore included as a feature in its own right,
and the linearity requirement is stated explicitly in Section~\ref{sec:protocol}. Similarly, the
ideality factor assumes that $\Voc$ varies logarithmically with illumination over the fitted
intensity range. The prediction \emph{targets}, in contrast, are SCAPS input parameters. The bulk
and interface defect densities are known by construction in simulation, whereas in a real device
they would have to be determined independently, for example by space-charge-limited-current or
admittance measurements.

The fifteen input features fall into three groups:
\begin{itemize}
\item \emph{Temperature series (six features):} the activation energy $\Ea$, obtained by linear
  extrapolation of $\Voc(T)$ to \SI{0}{\kelvin}; its deficit $\Ea - \Eg$ relative to the absorber
  band gap; $\mathrm{d}\Voc/\mathrm{d}T$ and the linearity of that fit;
  $\mathrm{d}\mathrm{FF}/\mathrm{d}T$; and the relative $\mathrm{d}\Jsc/\mathrm{d}T$.
\item \emph{Intensity series (five features):} the ideality factor obtained from
  $\mathrm{d}\Voc/\mathrm{d}\ln\Phi$, together with the linearity and curvature of that fit; the
  fill-factor slope $\mathrm{d}\mathrm{FF}/\mathrm{d}\ln\Phi$; and the ratio of $\Jsc$ per sun
  between the brightest and the dimmest illumination.
\item \emph{Standard test conditions (four features):} $\Voc$, $\Jsc$, FF and PCE.
\end{itemize}

\paragraph{Diagnostic classifier and metrics} The seven classes are predicted by independent
one-versus-rest models, so that a device can receive more than one label. Performance is
reported mainly as the macro $F_1$ score, defined as the unweighted mean of the per-class $F_1$
scores. Exact-match accuracy is defined as the fraction of devices whose predicted label set is
identical to the true label set.

\paragraph{Physics-rule baseline} \citet{kumar2021} describe the signatures that individual device
shortcomings leave in $\Voc(T)$ and $\Voc(\Phi)$, but they read these signatures qualitatively
from plots and do not specify numerical decision rules. A rule set codifying those signatures was
therefore implemented, with every threshold calibrated on the healthy devices of the training set
(5th and 95th percentiles) rather than chosen by hand, so that the comparison does not depend on
an arbitrary choice of cut-off values. An ideality factor above the healthy range indicates
parasitic resistance; an $\Ea$ below the healthy range indicates a contact barrier; a low $\Voc$
with a steep $\mathrm{d}\Voc/\mathrm{d}T$ indicates bulk recombination; a low $\Voc$ with a low
ideality factor indicates interface recombination; and a low FF at normal $\Voc$ indicates a
transport limitation. Because the rules cannot distinguish which interface is responsible, they
are credited with a correct answer if either interface class is true. In practice, this
concession has no effect: no device is predicted as exactly the two interface classes, and
scoring with and without it gives the same exact-match accuracy of 58.8\%.

Measurement noise was applied to the simulated $\Voc$, temperature, intensity, $\Jsc$ and FF
values before the features were computed, with magnitudes of \SI{3}{\milli\volt},
\SI{1}{\kelvin}, 5\%, 2\% and 1\%, respectively. In the robustness study
(Section~\ref{sec:requirements}), these magnitudes were scaled by a common multiplier.
\section{Results and discussion}

\subsection{Surrogate accuracy depends on how the parameter space is sampled}
\label{sec:sampling}

When trained on 75\% of the factorial grid and tested on the remaining 25\%, XGBoost predicts PCE
with $\Rtwo = 0.9997$ (RMSE 0.06\% absolute), the highest score among the five models on this
test. Both the ranking and the magnitude agree with previous SCAPS-based machine-learning
studies, which report 0.9998 for CsPbI\textsubscript{3} and 0.997 for
Sb\textsubscript{2}S\textsubscript{3} for the same model under the same 75/25 protocol
\citep{mustafa2025,ikram2026}. The present value is thus typical of the field, and the analysis
that follows concerns the evaluation protocol rather than any particular study. The same concern
applies to the one-factor-at-a-time sweeps that remain common in purely numerical studies of this
absorber \citep{amin2026}, which explore an even smaller part of the parameter space than a
factorial grid. When the same model is evaluated on 300 independent devices drawn at random from
the same parameter ranges, however, its $\Rtwo$ falls to 0.579, and its $\Jsc$ prediction fails
entirely ($\Rtwo = -0.34$).

Both values are obtained from a single model trained on a single dataset and differ only in the
location of the test devices. Table~\ref{tab:matched} lists the corresponding PCE results for all
five models, each fitted to the same 3,072 grid devices (the 75\% split used throughout this
section) and scored on both test sets. This design ensures that the decline in accuracy is not an
artifact of training-set size or of a comparison across experiments. The tree-based models, whose
predictions are piecewise constant, lose most of their accuracy off the grid, whereas the smooth
models retain it.

The test devices were drawn by Latin hypercube sampling on the natural scale of each parameter:
log-uniformly for the five density parameters, whose grid levels are spaced by decades, and
uniformly for the absorber thickness. Every sampled value lies within the parameter box of the
grid, so the comparison measures interpolation between grid nodes rather than extrapolation
beyond them. All 300 test simulations converged, as did all 4,096 grid simulations and all 3,000
space-filling simulations; consequently, no selection based on convergence affects the
comparison.

\begin{table}[tbp]
\centering
\caption{Coefficient of determination ($\Rtwo$) for PCE prediction by five models trained on the same 3,072 grid devices and scored on two test sets. The two columns differ only in the location of the test devices.}
\label{tab:matched}
\begin{tabular}{@{}lrr@{}}
\toprule
Model & Held-out grid (1,024) & Off-grid (300) \\
\midrule
XGBoost        & 0.9997 & \textbf{0.579} \\
Random forest  & 0.9987 & 0.811 \\
MLP            & 0.9979 & 0.994 \\
Quadratic      & 0.9689 & 0.955 \\
Linear         & 0.8024 & 0.725 \\
\bottomrule
\end{tabular}
\end{table}

\begin{figure}[tbp]
\centering
\includegraphics[width=0.92\textwidth]{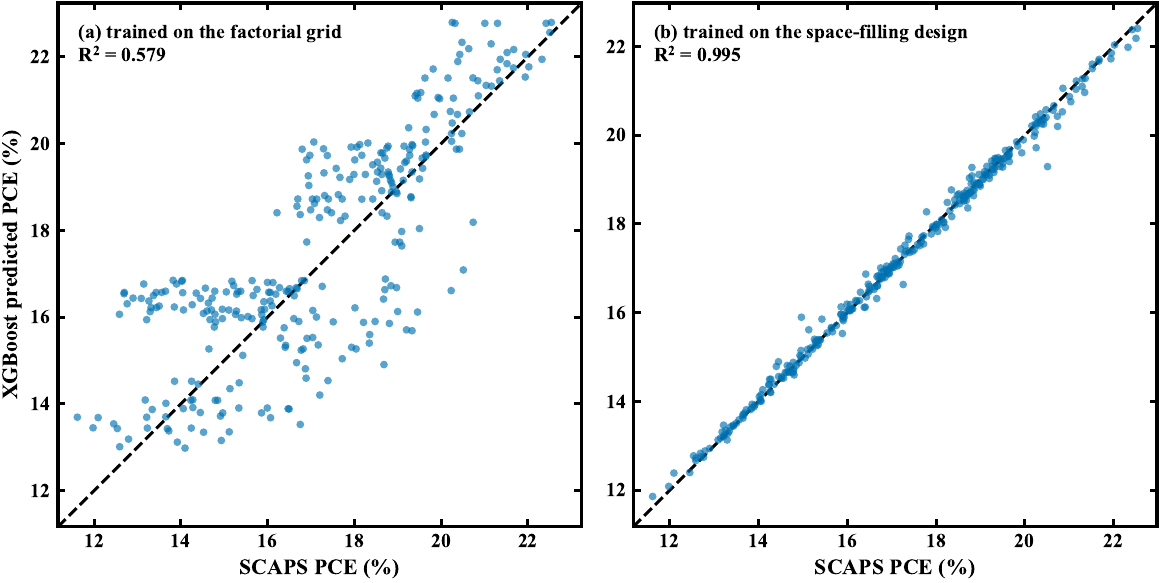}
\caption{Predicted against simulated PCE for the same 300 independent off-grid devices. XGBoost
was trained on (a) 3,072 devices of the factorial grid, which form the 75\% split of
Table~\ref{tab:matched}, and (b) the 3,000 devices of the space-filling design. The $\Rtwo$
values shown in the panels (0.579 and 0.995) are those reported in Table~\ref{tab:matched} and
Section~\ref{sec:sampling}. The horizontal banding in (a), which is absent in (b), reflects the
piecewise-constant predictions that a tree ensemble produces between grid levels.}
\label{fig:parity}
\end{figure}

The cause of this decline is structural. With four levels per input, a tree ensemble can place
split points only between grid values, so its predictions are piecewise constant across precisely
the regions that the grid does not sample \citep{numata2023}. Training the same model on the
3,000 space-filling devices, a slightly smaller training set than the 3,072 grid devices,
restores the off-grid accuracy to $\Rtwo = 0.995$, with a bootstrap 95\% confidence interval of
0.993--0.997 (Fig.~\ref{fig:parity}). This interval lies far from the grid-trained value of
0.579, so the contrast greatly exceeds the sampling uncertainty of the test set.

Because the two training sets differ in size as well as in design, a control experiment was
performed with 2,500 devices drawn from each set, using three independent draws per strategy. In
this experiment, grid-trained XGBoost reaches $\Rtwo = 0.582 \pm 0.002$ off the grid and
space-filling-trained XGBoost reaches $0.9940 \pm 0.0002$, close to the values obtained with the
full training sets (0.579 and 0.995 in Fig.~\ref{fig:parity}). The spread over draws is two
orders of magnitude smaller than the difference between the strategies, which confirms that the
effect arises from the sampling design rather than from the number of simulations.

An important practical consequence is that this deficiency cannot be detected from within the
grid. Five-fold cross-validation within the full grid ranks XGBoost first ($\Rtwo = 0.9998$),
followed by the random forest (0.9984) and the MLP (0.9981), whereas the same three models,
trained on the full grid and tested off the grid, are ranked in exactly the reverse order
($\Rtwo = 0.580$, 0.812 and 0.994, respectively). Cross-validation within the space-filling
training set, by contrast, reproduces the off-grid ranking (0.9969, 0.9947 and 0.9856 for the
MLP, XGBoost and random forest, respectively). A practitioner who samples on a grid and selects a
model by cross-validation will therefore tend to choose the model that generalizes worst between
the grid points.

The value of 0.579 thus characterizes a grid-trained surrogate on devices located between the
grid nodes; it is not the accuracy of the surrogates recommended in this work. When trained on a
space-filling design of comparable size, the same architectures reach 0.995 (XGBoost) and 0.997
(MLP) on the same independent devices. These values are comparable to those reported in the
literature for grid test points, but they are obtained here on an independent test set, which
grid sampling does not provide.

Two qualifications apply. First, the space-filling model can also be scored on the 4,096 grid
nodes, where XGBoost reaches $\Rtwo = 0.977$ and the MLP 0.992. Although these nodes are excluded
from its training, they form a structured lattice rather than an independent sample from the test
distribution, and this result is therefore not treated as a third measure of generalization.
Second, the 4,096 grid devices, the 3,000 space-filling devices and the 300 test devices were all
generated from the same definition file, with the same parameter mappings and the same fixed
device settings, and no configuration is shared between any training set and any test set. The
contrast therefore cannot be attributed to a difference in simulation setup.

All of the above results concern \emph{interpolation}. Genuine extrapolation reduces the accuracy
further, but only moderately: when the models are trained only on absorbers thinner than
\SI{0.6}{\micro\metre} and tested on thicker ones, XGBoost reaches $\Rtwo = 0.982$ and the MLP
0.950, compared with interpolation values of 0.990 and 0.996 in the same experiment.

With space-filling training, the model ranking also changes. The MLP becomes the most accurate
model for PCE ($\Rtwo = 0.9967$ on the held-out split and 0.9973 off the grid), followed by
XGBoost (0.9945 and 0.9950, respectively). The MLP also leads for $\Voc$ and FF, whereas XGBoost
remains the best model for $\Jsc$. Because the margins are small (at most 0.003 in $\Rtwo$), no
general superiority of one architecture is claimed; the grid-based ranking reported in the
literature reflects the test design rather than intrinsic properties of the models. The SHAP
attributions below are computed for the XGBoost model, for which exact tree-based attributions
are available.

As a consistency check on the four separately trained models, the product
$\Voc \times \Jsc \times \mathrm{FF}$ of the predicted values was compared with the separately
predicted PCE. The two agree to within 0.10\% absolute on average, with a maximum deviation of
0.68\%, whereas the corresponding SCAPS outputs are exactly consistent by construction.

\begin{figure}[tbp]
\centering
\includegraphics[width=0.46\textwidth]{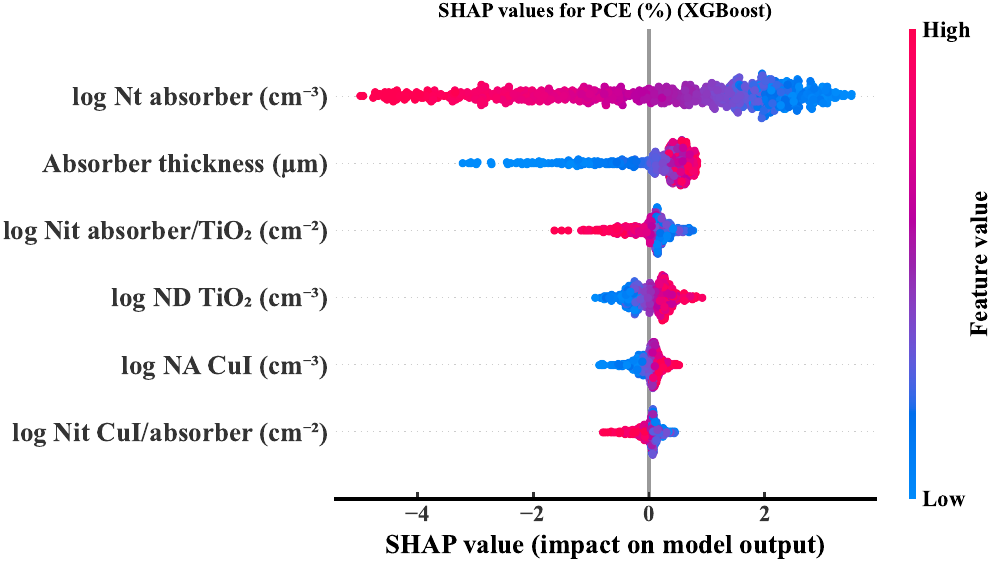}\hfill
\includegraphics[width=0.46\textwidth]{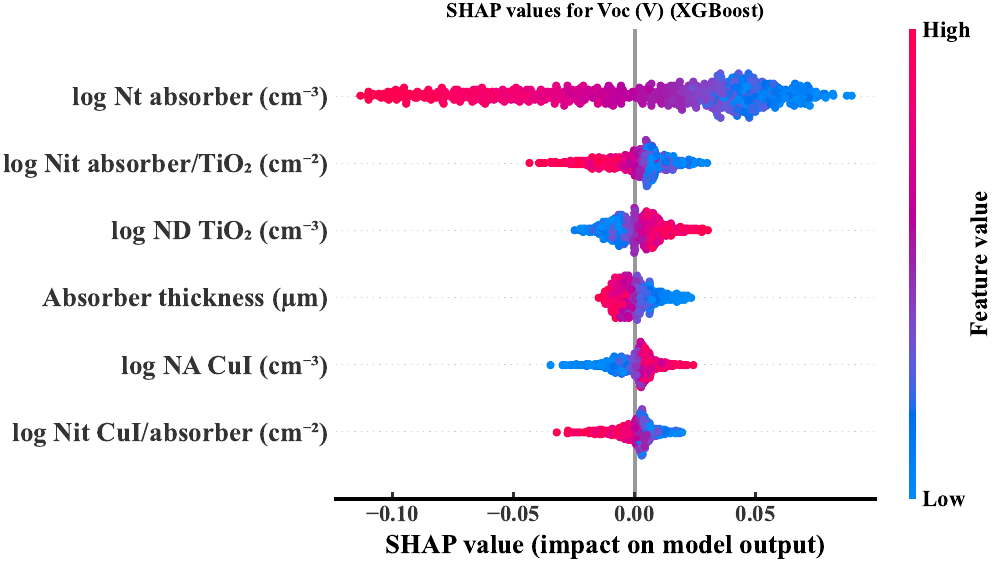}\\[4pt]
\includegraphics[width=0.46\textwidth]{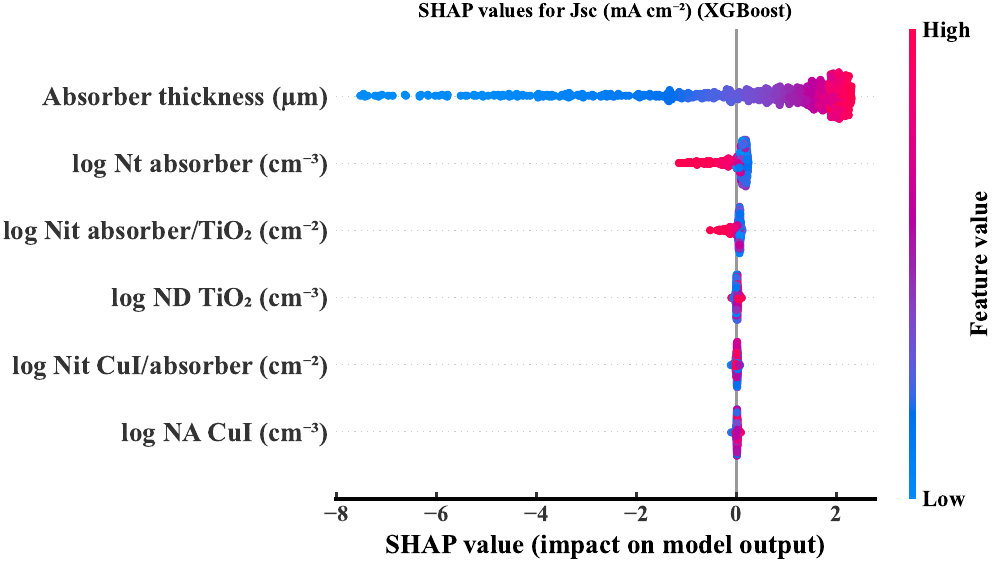}\hfill
\includegraphics[width=0.46\textwidth]{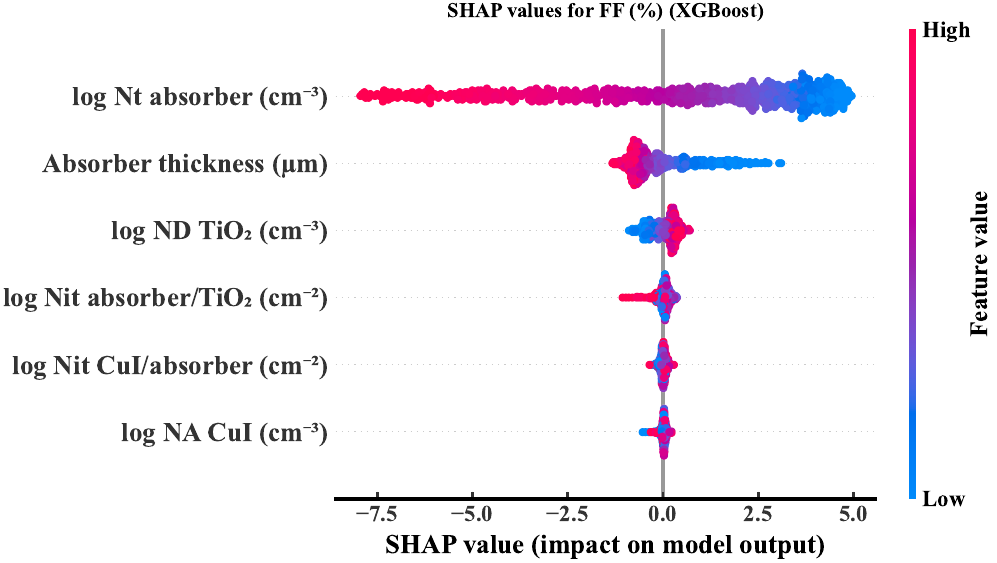}
\caption{SHAP attributions of the space-filling-trained XGBoost surrogate, computed on held-out
devices, for (a) PCE, (b) $\Voc$, (c) $\Jsc$ and (d) FF (left to right, top to bottom). Bulk
defect density dominates PCE and $\Voc$, and absorber thickness dominates $\Jsc$. FF is governed
jointly by the defect density and the transport-layer doping.}
\label{fig:shap_all}
\end{figure}

The leading SHAP ranking is qualitatively stable with respect to the sampling design, although
the attributed magnitudes change. Bulk defect density dominates PCE under both designs,
accounting for 47\% of the total SHAP magnitude with grid training and 57\% with space-filling
training, followed by absorber thickness with 25\% and 18\%, respectively; all values are
computed on held-out devices only (Fig.~\ref{fig:shap_all}). The ordering of the leading inputs
obtained from grid-trained models thus survives the change of design, even though their reported
accuracy does not. SHAP attributions quantify the contribution of each input to the \emph{model}
prediction; although they are consistent with the device physics in this case, they do not by
themselves constitute causal evidence.

The sampling result has implications beyond this device. Tree ensembles are the most widely used
regressors in SCAPS-based studies, and the factorial grid is the most widely used design. This
combination produces test points that coincide with the training levels, so that the reported
accuracy measures agreement at the grid nodes rather than the ability to interpolate between
them. The physical conclusions of such studies are not invalidated, and the SHAP rankings
obtained here are the same for both designs; the reported $\Rtwo$ values should not, however, be
interpreted as predictive accuracy for devices between the grid points. Space-filling designs
remove this problem, and in this work they also required fewer simulations in total (3,000
compared with 4,096).

\subsection{Forward optimization}
\label{sec:optimisation}

A surrogate trained on the space-filling set was used to screen 200,000 candidate devices, and
the highest-ranked candidate and four diverse runners-up were then re-simulated in SCAPS. All
five simulations completed successfully, and their results agree with the predictions to within
0.14 percentage points. The best SCAPS-verified candidate reaches an efficiency of 22.94\%
($\Voc = \SI{0.854}{\volt}$, $\Jsc = \SI{37.24}{\milli\ampere\per\centi\metre\squared}$,
$\mathrm{FF} = 72.1\%$), with an absorber thickness of \SI{0.74}{\micro\metre}, a bulk defect
density of \SI{1e14}{\per\centi\metre\cubed}, interface defect densities of approximately
\SI{1e10}{\per\centi\metre\squared} and doping densities of approximately
\SI{1e19}{\per\centi\metre\cubed} in both transport layers. For comparison, the untuned baseline
of the same definition file reaches 21.6\%.

Four qualifications bear on the interpretation of this value. First, it is the best candidate
that \emph{this search found and verified} rather than a global optimum, since 200,000 candidates
were ranked but only five were simulated. It also does not exceed the best of the 4,096 grid points,
which reaches 22.95\%; the search matches that value from a different region of the parameter
space rather than improving on it. Second, the surrogate is considerably less accurate in the
region selected by the search: on independent devices, its $\Rtwo$ for PCE is 0.995 overall but
only 0.786 within the top decile of efficiency. Moreover, selecting the maximum of noisy
predictions favors points at which the model over-predicts, which explains why all five verified
candidates fall below their predicted values. Third, the optimum is a plateau rather than a
single point. Repeating the search with different candidate pools and model seeds gives a stable
predicted efficiency of $23.03 \pm 0.04\%$ over six repeats but markedly different parameter
values, with the absorber thickness, for example, ranging from
\SIrange{0.64}{0.91}{\micro\metre}. The individual optimized parameters should therefore not be
over-interpreted. Because the search also operates where the surrogate is least accurate, this
part of the study is affected, in milder form, by the sampling problem identified in
Section~\ref{sec:sampling}.

Fourth, the search is limited in scope. The six searched parameters are all properties of a fixed
TiO\textsubscript{2}/\allowbreak\absorber{}/\allowbreak CuI stack, and neither the transport materials nor the
back-contact metal were varied. \citet{amin2026} report 26.01\% for the same absorber by changing
precisely these fixed choices, replacing TiO\textsubscript{2} with
In\textsubscript{2}Se\textsubscript{3} and Au with Ni and tuning the back-contact work function,
a gain larger than any improvement available within the present parameter space. The value
reported here is therefore not a performance ceiling for \absorber{}, nor is it presented as one; it
is the efficiency that a surrogate search recovers within one architecture whose parameters are
restricted to literature-supported ranges. The contribution of this section lies in the search
protocol and the associated assessment of accuracy rather than in the peak efficiency.

Three of the six optimized parameters lie at a boundary of the search space: the bulk defect
density and the ETL-side interface defect density are at their minima, and the
TiO\textsubscript{2} doping density is at its maximum. The design space therefore contains no
interior maximum, and the contribution of the surrogate in this case is accurate interpolation
rather than the discovery of an unexpected optimum. Moreover, the defect densities are
material-quality parameters rather than quantities that an experimenter sets directly, so this
configuration should be regarded as a performance target rather than as a fabrication recipe.
\subsection{Diagnosis of degradation mechanisms}
\label{sec:diagnosis}

\begin{figure}[tbp]
\centering
\includegraphics[width=\textwidth]{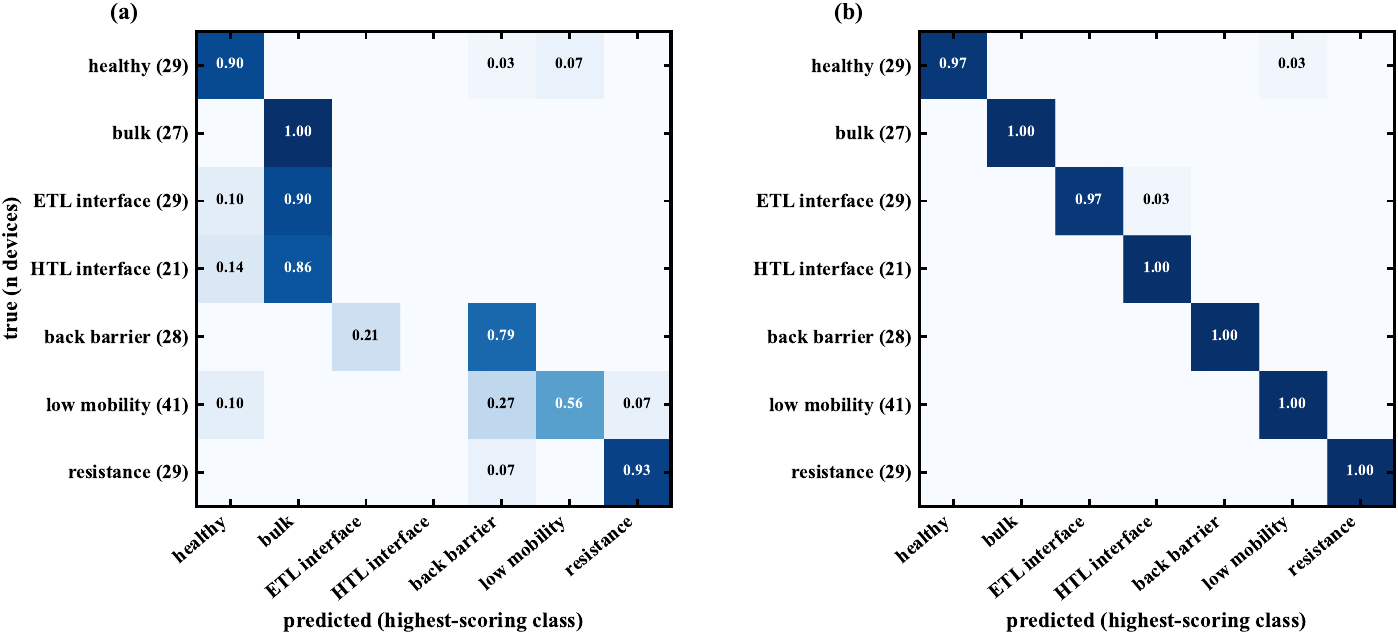}
\caption{Confusion matrices for the single-degradation devices. Each method was required to name
one class instead of giving its native multi-label output: (a) calibrated physics rules, with a
forced-choice accuracy of 61\%; (b) XGBoost classifier, with 99\%. Row labels give the number of
test devices in each class.}
\label{fig:confusion}
\end{figure}

On 298 independent test devices, the XGBoost classifier recovers the introduced degradation
mechanisms from $\Voc(T)$ and $\Voc(\Phi)$ with macro $F_1 = 0.938$, compared with 0.687 for the
calibrated physics rules. Five-fold cross-validation on the training set gives
$0.930 \pm 0.010$, indicating that the held-out score does not depend on the particular split.
For the 204 single-degradation devices, each method was additionally required to name its single
highest-scoring class. Under this forced-choice scoring, the accuracy is 99\% for the model and
61\% for the rules (Fig.~\ref{fig:confusion}).

Although the forced choice is applied identically to both methods, it does not reflect how
either method natively reports its output; the rules, for example, name more than one mechanism
for 49 of the 204 devices. Two further scoring conventions were therefore considered. The first
counts a prediction as correct if the true class appears anywhere in the predicted set, a
convention that favors a method naming several mechanisms; under it, the rules reach 82.4\% and
the model 97.5\%. The second requires exact agreement between the predicted and true label sets
and gives 58.8\% for the rules and 95.1\% for the model. The model thus leads under every
convention, by margins of 15 to 38 percentage points. The rules fail systematically: between 86\%
and 90\% of the interface-limited devices are attributed to bulk recombination, because both
mechanisms lower $\Voc$ and steepen $\mathrm{d}\Voc/\mathrm{d}T$.

The advantage of the model is largest when mechanisms coexist. For devices with two simultaneous
degradations, the rules recover the complete label set in 22\% of cases and the model in 69\%.
Among the learned models, XGBoost (0.938) is followed by the random forest (0.928) and the MLP
(0.915), whereas logistic regression (0.698) performs only slightly better than the rules,
indicating that the task is not linearly separable in these features. The rule baseline is also
close to its own optimum: recalibrating its thresholds at the 1st, 5th, 10th and 25th percentiles
of the healthy population gives macro $F_1$ values between 0.539 and 0.692, so the comparison is
not made against an artificially weak baseline.

Because the seven classes are predicted by independent one-versus-rest models, a device could in
principle be labeled both healthy and defective, or receive no label at all. Neither case is
frequent: no test device received a contradictory label, and the three of the 298 devices that
received no label are reported as indeterminate rather than assigned to a class. Tuning a
separate decision threshold for each class by cross-validation on the training set did not
improve the result (macro $F_1 = 0.938$ in both cases, and 0.768 and 0.775 on mixed-defect
devices), so the default threshold of 0.5 was retained. Finally, the test devices are no closer
to their nearest training device than the training devices are to one another, with median
nearest-neighbor distances in standardized feature space of 0.342 and 0.335, respectively. This
result confirms that the independent sampling of the test set introduces no leakage through
near-duplicate devices.

The severity of the bulk defect density is predicted well, with $\Rtwo = 0.92$ and a mean
absolute error (MAE) of 0.08 decades (Fig.~\ref{fig:severity_regression}), whereas the interface
defect densities are predicted poorly ($\Rtwo = 0.53$ and 0.02 for the ETL and HTL sides,
respectively). These values are obtained on
devices that the classifier itself identifies as bulk-limited, rather than on devices whose true
class is assumed to be known. The classifier identified 61 such devices in the test set, all of
them correctly, so the end-to-end pipeline attains the same accuracy as the case with known
labels. The weakness for interfaces reflects what these measurements can determine rather than a
limitation of the regressor: as the controlled sweeps described below show, the interface defect
density is the least identifiable of the six mechanisms from $\Voc(T)$ and $\Voc(\Phi)$, because
it changes the measurable signals much less than the bulk defect density does.

\begin{figure}[tbp]
\centering
\includegraphics[width=0.62\textwidth]{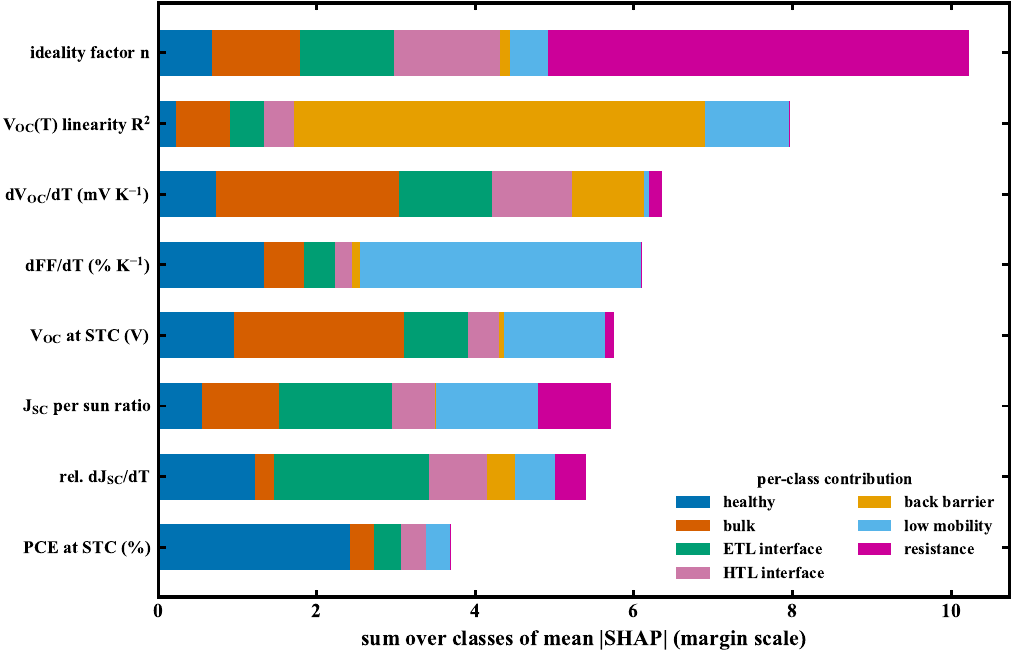}
\caption{Per-class SHAP attributions of the diagnostic classifier. The ideality factor dominates
the resistance class, and the linearity of $\Voc(T)$ dominates the contact-barrier class.
$\mathrm{d}\Voc/\mathrm{d}T$, together with $\Voc$, dominates the bulk class.}
\label{fig:shap2}
\end{figure}

\begin{figure}[tbp]
\centering
\includegraphics[width=0.40\textwidth]{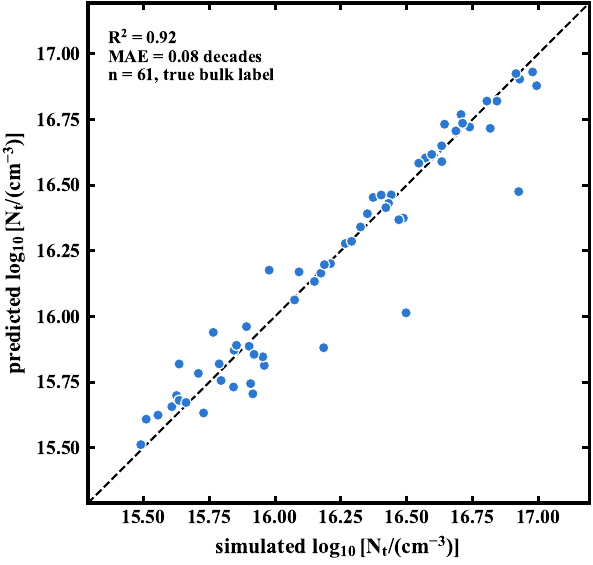}
\caption{Predicted against simulated bulk defect density for the test devices that contain a bulk
defect. The dashed line indicates the identity.}
\label{fig:severity_regression}
\end{figure}

The systematic misattribution by the rules has a specific cause. The signal that distinguishes
interface from bulk recombination, a lower ideality factor combined with an activation-energy
deficit, is too weak for a threshold on any single feature to separate the two mechanisms. The
learned model, in contrast, uses the features jointly, and its per-class SHAP rankings recover
the signatures that the experimental literature associates with each mechanism
\citep{hegedus2004,tress2018,kumar2021} (Fig.~\ref{fig:shap2}): the ideality factor dominates the
resistance class, the linearity of $\Voc(T)$ the contact-barrier class, the temperature slope of
FF the transport class, and $\mathrm{d}\Voc/\mathrm{d}T$ together with $\Voc$ the bulk class.

\begin{figure}[tbp]
\centering
\includegraphics[width=0.65\textwidth]{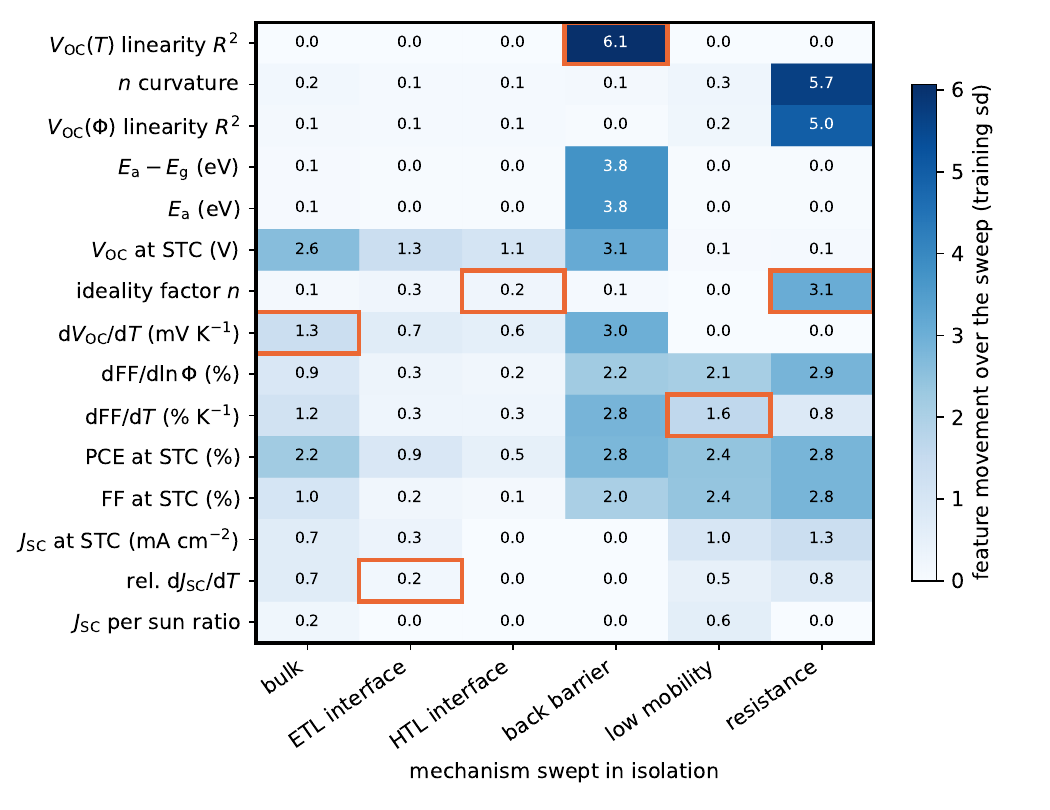}
\caption{Simulated response of each feature when a single mechanism is swept in isolation. Each
cell gives the peak-to-peak variation of the feature over the sweep (its maximum minus its
minimum across the sweep levels), divided by the standard deviation of the feature across the
training set so that features with different units can be compared. Orange
outlines mark the feature on which the classifier relies most for each mechanism (highest mean
$|$SHAP$|$). The back-barrier and resistance signatures are sharp and exclusive, whereas the
two interface columns are uniformly weak, which explains why these classes are the most
difficult to diagnose.}
\label{fig:sweep}
\end{figure}

SHAP, however, explains the fitted model and distributes credit among correlated inputs, so a
high attribution may indicate that a feature is a convenient proxy rather than a quantity that
the mechanism actually changes. The attributions were therefore tested directly against the
simulator. Each mechanism was swept individually, with its defining parameter varied from the
middle of the healthy range to the far end of the degraded range while every other parameter was
held at its healthy midpoint. The response of each feature was quantified as its peak-to-peak
variation over the sweep, $\max_i x_i - \min_i x_i$, where $i$ denotes the sweep level, divided
by the standard deviation of that feature across the training set so that features with
different units can be compared (Fig.~\ref{fig:sweep}). The sweep comprises 60 devices, each
simulated at nine conditions.

The resulting signatures are distinct and largely specific to each mechanism. Sweeping the
back-contact work function changes the linearity of $\Voc(T)$ by 6.1 standard deviations and the
extrapolated $\Ea$ by 3.8, while both features remain essentially unchanged for every other
mechanism ($\le 0.1$). Sweeping the resistances changes the curvature of $\Voc(\Phi)$ by 5.7, the
linearity of that fit by 5.0 and the ideality factor by 3.1, again with no response to the other
mechanisms. For four of the six mechanisms (bulk defects, back barrier, low mobility and
parasitic resistance), the feature on which the model relies most responds to that mechanism in
isolation, by 1.4 to 6.1 standard deviations. The physical interpretation of the SHAP results is
therefore supported by the simulator itself and not only by analogy with the literature.

The two interface classes behave differently, and the difference is informative. Sweeping either
interface defect density from healthy to fully degraded changes no feature by more than 1.3
standard deviations, compared with a maximum of 2.4--6.1 for the other four mechanisms, and the
features on which the model relies for the interface classes change by only 0.2. An interface
defect thus leaves a much weaker signature in these measurements than a contact barrier or a
series resistance does. This is an identifiability limit of the measurement set rather than a
shortcoming of the classifier: no model, whether learned or rule-based, can separate mechanisms
that the measurements barely distinguish, and when the signal is this weak, the attribution is
distributed over residual correlations between features.

This conclusion is consistent with three independent observations reported elsewhere in this
work. The HTL interface has the lowest per-class $F_1$ (0.72) and the lowest area under the ROC
curve (0.912), and it is also the least well-calibrated class (Section~\ref{sec:measured}). In
addition, restoring an interface label recovers the least efficiency of any class when mechanisms
coexist (Section~\ref{sec:datasets}). Together with the sweep, these four results indicate that
interface recombination is the mechanism least well constrained by $\Voc(T)$ and $\Voc(\Phi)$.
Improving its diagnosis requires an additional measurement channel, such as admittance
spectroscopy, capacitance--voltage profiling or spectrally resolved quantum efficiency, rather
than a better classifier operating on the same inputs.

The scope of the sweep is limited. It shows how the features respond when one parameter is
varied in isolation in this device; it does not convert SHAP into a causal attribution, and the
features remain mutually correlated in the training datasets. The model therefore does not
replace physical analysis but applies the same physical signals without the simplification
imposed by fixed thresholds.

The advantage of the model over the rules increases with device complexity, which is the
experimentally most relevant regime, because a fabricated cell is rarely subject to a single loss
mechanism. As shown in Section~\ref{sec:combinations}, however, this advantage has been measured
only on combinations of mechanisms that are present in the training set.

\subsection{Generalization to unseen combinations of mechanisms}
\label{sec:combinations}

The multi-mechanism accuracy reported above is measured on devices whose combination of
mechanisms also occurs in the training set. All fifteen pairs of the six degradation mechanisms
are represented in training, with 27 to 47 devices each, so the exact-match accuracy of 69\% is
that of a model trained on every pair. A separate question is whether the model \emph{composes}
individual mechanisms or merely recognizes familiar combinations, which is decisive for a device
that exhibits a combination of mechanisms absent from the training set.

This question was addressed within the training set, leaving the independent test set unchanged.
For each of the fifteen pairs, the devices containing that pair were divided into two halves, and
two models were scored on the same evaluation half: one trained with the other half included, so
that the combination was familiar to it, and one trained after every device containing that pair
had been removed.

\begin{figure}[tbp]
\centering
\includegraphics[width=0.48\textwidth]{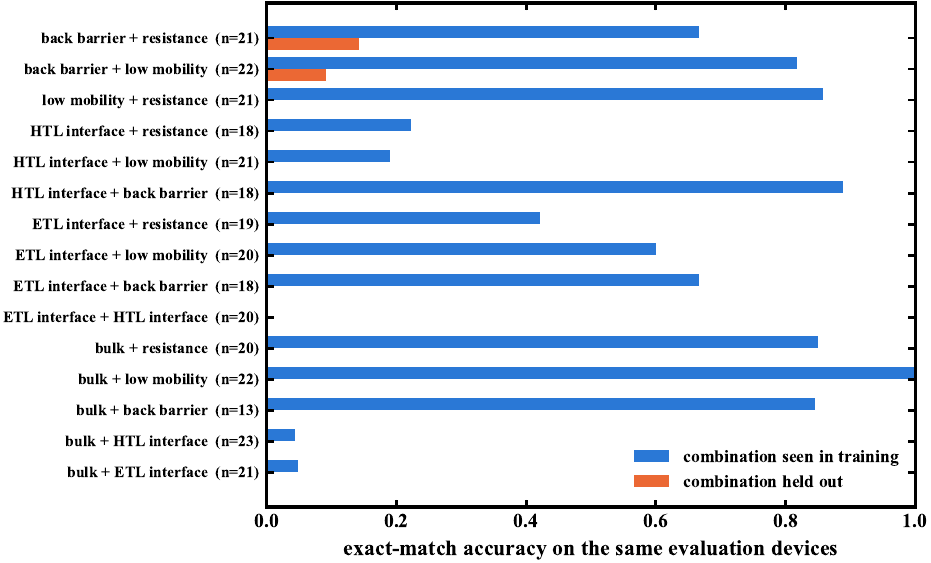}
\hfill
\includegraphics[width=0.48\textwidth]{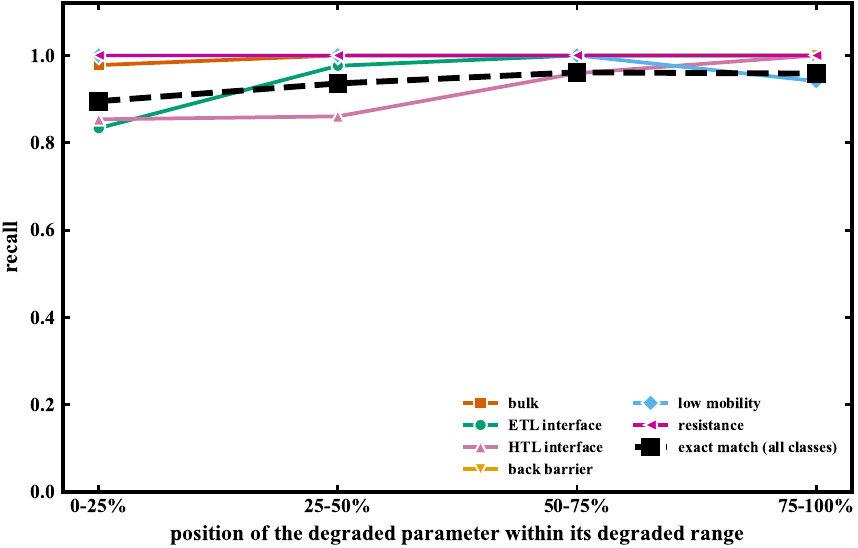}
\caption{Generalization of the diagnostic model. (a) Exact-match accuracy on the same evaluation
devices when the pair of mechanisms they contain is present in training (blue) and when every
device containing that pair is removed from training (orange). (b) Per-class recall and
exact-match accuracy as a function of how far the degraded parameter lies within its degraded
range, obtained from out-of-fold predictions within the training set.}
\label{fig:generalisation}
\end{figure}

The difference between the two models is large. Exact-match accuracy on the same evaluation
devices falls from 0.541 when the combination is familiar to 0.016 when it is new
(Fig.~\ref{fig:generalisation}a). The model therefore does not generalize to unseen combinations
of mechanisms.

The failure is systematic rather than random, in that the classifier typically recovers one
constituent mechanism and misses the other. The mean recall over the two constituent mechanisms
is close to 0.5 for every pair, with one mechanism usually reported with a recall above 0.9 and
the other almost entirely missed. Although this behavior does not make the output usable as a
complete diagnosis, it shows that the error is a predictable omission rather than an arbitrary
misassignment.

Which mechanism is reported can be predicted from the sweep described in
Section~\ref{sec:diagnosis}. In all fifteen pairs, the reported mechanism is the one with the larger
feature signature in the sweep. This holds even for the closest pair, the back barrier (6.1
standard deviations) and resistance (5.7), which is also the only pair in which both mechanisms
retain partial recall. A quantity obtained independently from the simulator therefore predicts,
for each pair, which mechanism the model will report when it cannot report both.

Each pair is present in the training set but accounts for only about 2\% of it, which suggests
assigning a larger weight to multi-mechanism devices during training. Several weighting schemes
were tested: two-mechanism devices were upweighted by factors of 2, 4 and 8, the fifteen pairs
were reweighted to equal mass, and the combined two-mechanism mass was matched to the
single-mechanism mass. Across these schemes, the exact-match accuracy on multi-mechanism test
devices changed from 0.691 to values between 0.681 and 0.713. The best scheme, a uniform factor
of 4, changed the outcome for four of the 94 evaluation devices, three of which became correctly
classified and one misclassified, and the paired bootstrap interval for the change ($-0.021$ to
$+0.064$) contains zero. Reweighting therefore yields no significant improvement.

Class imbalance is thus not the factor that limits multi-mechanism recovery. This finding is
consistent with the identifiability interpretation given above: the model does not lack examples
of these pairings, but the measurements contain too little signal to distinguish the weaker
constituent. Consequently, every pairing that a diagnosis must cover has to be simulated
explicitly.

By contrast, performance depends only weakly on the severity of the degradation.
Single-degradation devices were grouped into quarters according to the position of the degraded
parameter within its degraded range and scored with out-of-fold predictions. From the mildest to
the most severe quarter, the exact-match accuracy is 0.895, 0.936, 0.961 and 0.958
(Fig.~\ref{fig:generalisation}b), and no severity threshold was found below which the diagnosis
fails. The only class that is appreciably weaker at low severity is again the HTL interface, whose
recall rises from 0.85 to 1.00 across the range.

\subsection{Requirements for reliable diagnosis}
\label{sec:requirements}

\paragraph{Measurement noise must be included during training} Noise is applied to every recorded
quantity before the features are computed: additive Gaussian noise for $\Voc$
(\SI{3}{\milli\volt}) and temperature (\SI{1}{\kelvin}), and multiplicative lognormal noise for
illumination (5\%), $\Jsc$ (2\%) and FF (1\%). The efficiency is then recomputed from the
perturbed values, so that each simulated operating point remains internally consistent. Each
noise level is evaluated with three independent measurement realizations on the same devices and
labels, so that differences between levels reflect the noise rather than a change of dataset,
and the mean and standard deviation over the three realizations are reported. The noise model
assumes independent Gaussian and lognormal errors. Hysteresis, preconditioning history and drift
during a temperature series are not represented, although all three affect real measurement
campaigns.

\begin{figure}[tbp]
\centering
\includegraphics[width=0.46\textwidth]{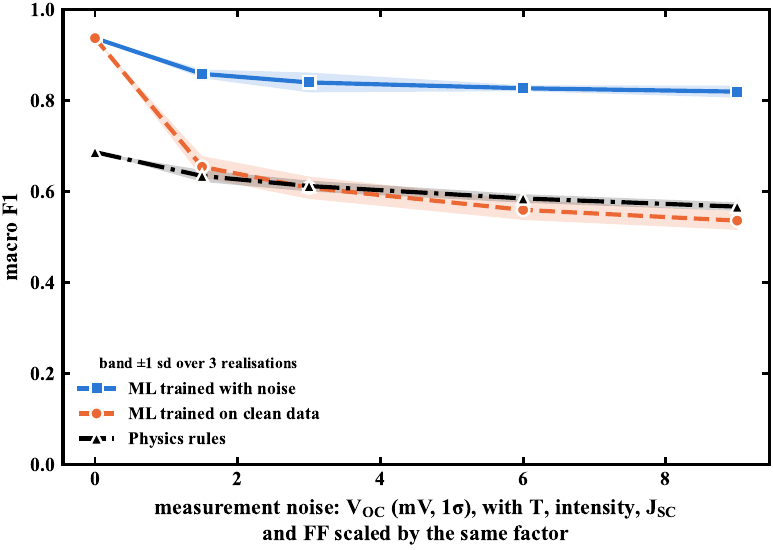}
\hfill
\includegraphics[width=0.5\textwidth]{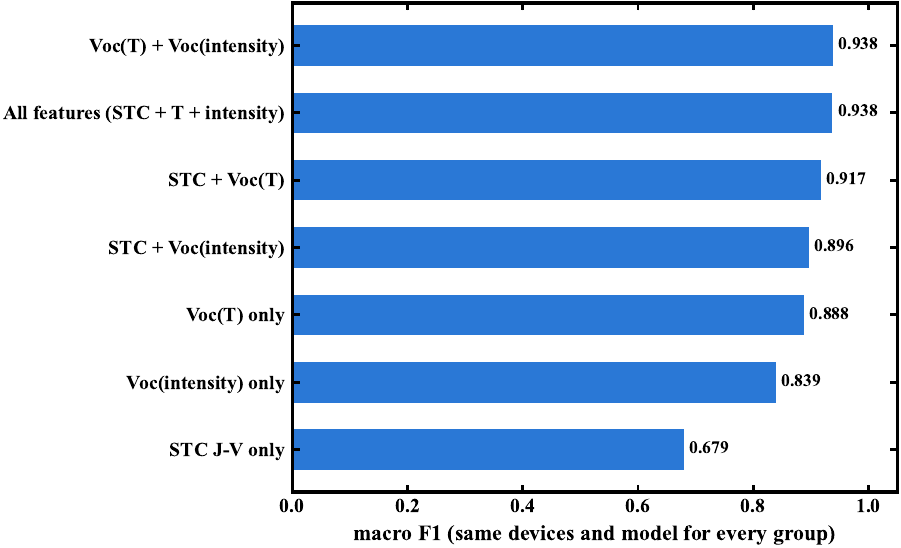}
\caption{(a) Macro $F_1$ as a function of measurement noise. Shaded bands indicate $\pm 1$
standard deviation over three independent measurement realizations per level. (b) Feature
ablation. The three lowest bars isolate each measurement series. The standard-condition rows
quantify the contribution of each series relative to a single $J$--$V$ curve.}
\label{fig:noise_ablation}
\end{figure}

A model trained on noise-free data performs poorly on noisy data, with its macro $F_1$ falling
from 0.938 to $0.609 \pm 0.025$ at the $1\times$ noise level, below the physics rules. When the
model is trained on noisy data and tested on independently generated noisy data, it retains
$0.859 \pm 0.010$ at $0.5\times$ and $0.820 \pm 0.013$ at $3\times$, the latter corresponding to
\SI{9}{\milli\volt} of $\Voc$ noise and 15\% intensity noise. Over the same range, the rule
baseline falls from 0.687 to 0.567 (Fig.~\ref{fig:noise_ablation}a). The $1\times$ level
corresponds to a well-maintained measurement setup. The $3\times$ level, which corresponds to a
thermocouple reading that deviates by \SI{3}{\kelvin} from the device temperature combined with
uncalibrated neutral-density filters, represents a pessimistic bound rather than a typical
measurement.

The errors considered above are independent between measurements. Calibration errors, by
contrast, are systematic, because a single thermocouple offset and a single filter transmission
factor apply to an entire measurement series. Adding a per-device systematic offset to the
$1\times$ random noise does not degrade the diagnosis: the macro $F_1$ changes from
$0.840 \pm 0.022$ to $0.846 \pm 0.007$ for an offset of \SI{1}{\kelvin}/3\% and is
$0.848 \pm 0.012$ for \SI{3}{\kelvin}/10\%. This behavior follows from the form of the features. A
constant intensity scale factor shifts $\ln\Phi$ by a constant and leaves the ideality factor,
which is a slope, unchanged, and a constant temperature offset $\delta$ shifts the extrapolated
$\Ea$ by only $(\mathrm{d}\Voc/\mathrm{d}T)\,\delta \approx \SI{5}{\milli\electronvolt}$ at
$\delta = \SI{3}{\kelvin}$. Random scatter, which degrades the fits themselves, is the more
demanding case, and the noise levels reported above are therefore the conservative ones.

\paragraph{Temperature and intensity series carry most of the information} The standard 1-sun
$J$--$V$ parameters alone give macro $F_1 = 0.679$. Adding the intensity series raises this
value to 0.896, adding the temperature series raises it to 0.917, and adding both raises it to
0.938, an improvement of 0.26 over a single $J$--$V$ curve (Fig.~\ref{fig:noise_ablation}b).
Without the standard-condition values, the intensity and temperature series give 0.839 and 0.888,
respectively, and 0.938 together. The standard-condition parameters are therefore informative on
their own but become redundant once both series are available. Five-fold cross-validation within
the training set gives the same ordering (0.680, 0.889, 0.904 and 0.930); the feature groups were
compared on this basis, and the test set was used only to report the final values.

\paragraph{Generalization across absorber thickness} When trained on absorbers thinner than
\SI{0.7}{\micro\metre} and tested on thicker ones, the model scores 0.913, compared with
$0.904 \pm 0.013$ for a reference model trained on the same number of devices drawn from the full
thickness range; extrapolation to thicker absorbers therefore incurs no loss of accuracy in this
case. In the reverse direction, the model scores 0.879, compared with $0.933 \pm 0.012$ for a
size-matched reference. This difference of 0.054 is a genuine loss of accuracy, which arises
because training on thick absorbers does not capture the collection-limited behavior of thin
devices. In both directions, the training and test sets are disjoint in thickness and contain
every class. The test addresses distribution shift in thickness only and provides no information
about unseen ranges of defect severity, other device architectures or other measurement
conditions.
\subsection{Physics-derived features outperform the tested end-to-end models}

The features defined in Section~\ref{sec:features} encode physical knowledge in the form of an
extrapolation to \SI{0}{\kelvin}, a logarithmic slope and a set of ratios. An alternative
approach is to supply the measured curves directly as model inputs. A direct comparison of the
two approaches is, however, confounded, because the physics features are derived from $\Voc$,
$\Jsc$ and FF at every temperature and every intensity, whereas a raw representation limited to
the two $\Voc$ curves carries strictly less information. Two raw representations were therefore
evaluated: the two $\Voc$ curves together with the standard-condition point (13 values), and an
information-matched set comprising $\Voc$, $\Jsc$ and FF at all five temperatures and all five
intensities (33 values). Each representation was evaluated with several architectures, so that
the effect of the representation could be separated from that of the model. All models were
trained and scored on the same 1,992 training devices and 298 test devices, matched by device
identifier rather than by row order, and every value is the mean over three random seeds
(Table~\ref{tab:cnn}).

\begin{table}[tbp]
\centering
\caption{Macro $F_1$ for combinations of input representation and model architecture, all evaluated on the same devices. Values are means over three seeds $\pm$ one standard deviation.}
\label{tab:cnn}
\begin{tabular}{@{}llr@{}}
\toprule
Input & Model & macro $F_1$ \\
\midrule
15 physics features (Section~\ref{sec:features}) & XGBoost & \textbf{0.935 $\pm$ 0.001} \\
15 physics features & MLP & 0.915 $\pm$ 0.007 \\
raw $\Voc$ curves only (13 values) & XGBoost & 0.811 $\pm$ 0.003 \\
raw $\Voc$ curves only (13 values) & 1-D CNN & 0.857 $\pm$ 0.007 \\
information-matched raw curves (33) & XGBoost & 0.855 $\pm$ 0.005 \\
information-matched raw curves (33) & MLP & 0.863 $\pm$ 0.012 \\
information-matched raw curves (33) & 1-D CNN & 0.848 $\pm$ 0.009 \\
raw curves + physics features & 1-D CNN & 0.880 $\pm$ 0.004 \\
\bottomrule
\end{tabular}
\end{table}

Three conclusions follow from Table~\ref{tab:cnn}. First, providing the raw-input models with the
information-matched measurements improves them, with the XGBoost score rising from 0.811 to
0.855, so part of the original gap reflects a difference in information content rather than in
representation. Second, the remaining gap is not caused by a weak convolutional network: a
standard MLP on the same 33 raw values reaches 0.863, comparable to the 0.848 of the
convolutional network, which indicates that the ceiling for raw inputs is a property of the
representation rather than of a particular architecture. Third, the effect of the architecture
is small compared with that of the representation. On the physics features, XGBoost leads the
MLP by 0.020, and on the raw values, XGBoost and the convolutional network differ by less than
0.010, whereas the gap between the best physics-feature model and the best raw-input model is
0.072. Supplying the physics features alongside the raw curves recovers much of this difference
(0.880), suggesting that the features provide a useful inductive bias rather than additional
information.

\begin{figure}[tbp]
\centering
\includegraphics[width=0.50\textwidth]{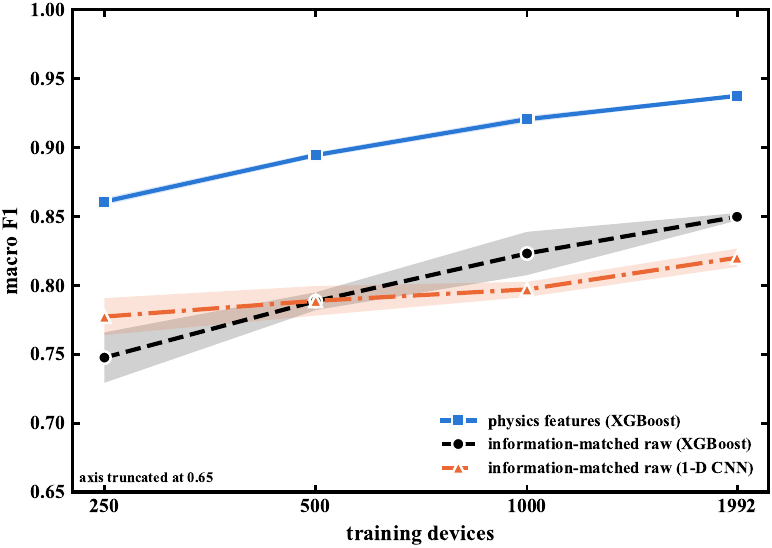}
\caption{Learning curves of the feature-based model and the convolutional network on raw curves.}
\label{fig:lc}
\end{figure}

The gap does not close as the training set grows within the range tested. The training subsets
used for the learning curves are nested, each being a prefix of a single shuffled device order,
so that a larger subset always contains the smaller ones. The feature-based model outperforms the
convolutional network on raw curves at every training-set size tested (Fig.~\ref{fig:lc}); the
physics features are therefore more data-efficient and retain their advantage up to the largest
dataset generated in this work, although whether the gap closes for larger datasets was not
tested. Nor is the gap an artifact of undertraining: the network is trained with early stopping
on a validation split of the training devices, so its stopping point adapts to the amount of
available data, and the test set is never used for this choice.

This conclusion is restricted to the conditions tested. For this simulated device distribution,
these measurement sets and the dataset sizes available here, physically motivated descriptors
outperform the tested raw-input and end-to-end architectures, which supports their use at dataset
sizes realistic for device simulation. Feature engineering is not claimed to be inherently
superior to representation learning, since a larger network, a different architecture or a
dataset several orders of magnitude larger might close the gap. Because only a limited set of
neural architectures was tested, the result describes the approaches examined here rather than
the best achievable performance.

\begin{figure}[tbp]
\centering
\includegraphics[width=0.50\textwidth]{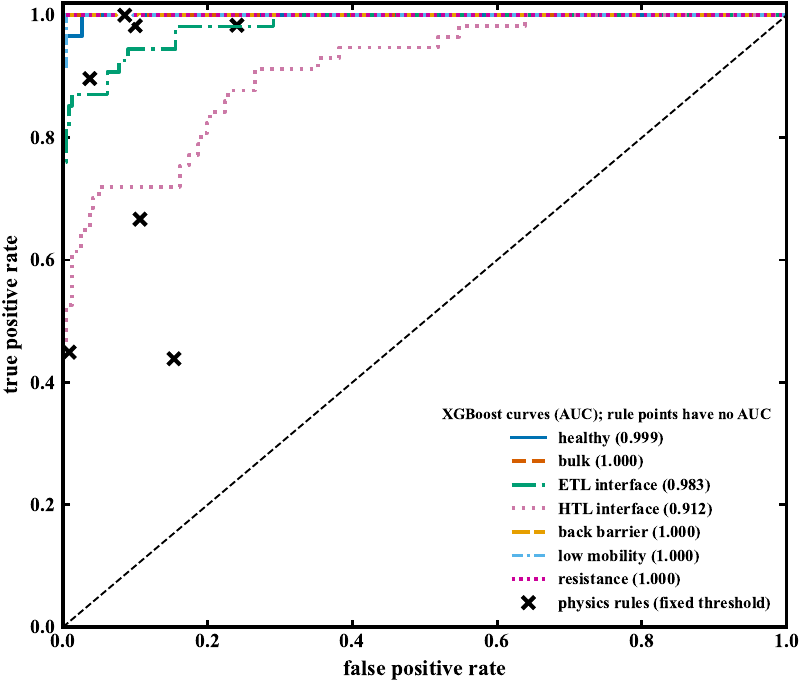}
\caption{Per-class ROC curves of the diagnostic classifier. The operating points of the
calibrated physics rules are marked.}
\label{fig:roc}
\end{figure}

Per-class discrimination is strong for every mechanism except the HTL-side interface. The areas
under the receiver operating characteristic (ROC) curve, rounded to three decimal places, are
1.000 for bulk defects, the back-contact barrier, low mobility and parasitic resistance, 0.999
for healthy devices, 0.983 for the ETL-side interface and 0.912 for the HTL-side interface,
giving a macro-averaged area of 0.985. In every case, the operating points of the physics rules
lie well below the ROC curves of the model (Fig.~\ref{fig:roc}).

\subsection{Exploratory application to measured devices}
\label{sec:measured}

This section addresses a narrower question than the rest of the paper: whether the required
features can be extracted from a \emph{published} measurement, and what a model trained only on
simulated \absorber{} devices predicts for a different absorber. The analysis is an exploratory
external transfer. With two devices and no verified class labels, it cannot measure experimental
accuracy and can establish only the validity of the feature extraction and the qualitative
consistency of the result with the attribution given by the original authors.

Two FASnI\textsubscript{3} cells with light-intensity-dependent $\Voc$ measurements
\citep[Fig.~S21b]{liu2020} were digitized. The measured curves are sampled at 14 intensities
spaced linearly from \SIrange{6.7}{100}{\milli\watt\per\centi\metre\squared}, whereas the
simulated devices span 0.01--1 sun. The features of both were therefore computed at the
\emph{same three intensities} (0.1, 0.3 and 1 sun), with the measured $\Voc$ interpolated in
$\ln\Phi$, so that no extrapolation is involved. The ideality factors obtained in this way, 1.91
and 1.57, agree with the published values of 1.98 and 1.62, both evaluated at the temperature of
\SI{298}{\kelvin} stated by the authors, which validates the feature extraction itself. All twelve
measured feature values lie within the range spanned by the simulated training set.

When restricted to band-gap-relative and dimensionless features, the model reaches macro
$F_1 = 0.907$ on held-out \emph{simulated} devices. This value depends on the composition of the
test set: when the model is retrained on the original and hard datasets, it scores 0.906 on the
ordinary test devices but only 0.700 on the deliberately harder ones, for a pooled value of
0.869. Pooled values should therefore be interpreted together with the composition of the test
set. Per-class performance is uneven, with $F_1$ scores of 0.98 for bulk defects, resistance and
low mobility, 0.93 for the back barrier and only 0.65 for the HTL interface, whose scores are
also the least well calibrated (largest bin-wise gap of 0.19).

\begin{figure}[tbp]
\centering
\includegraphics[width=0.92\textwidth]{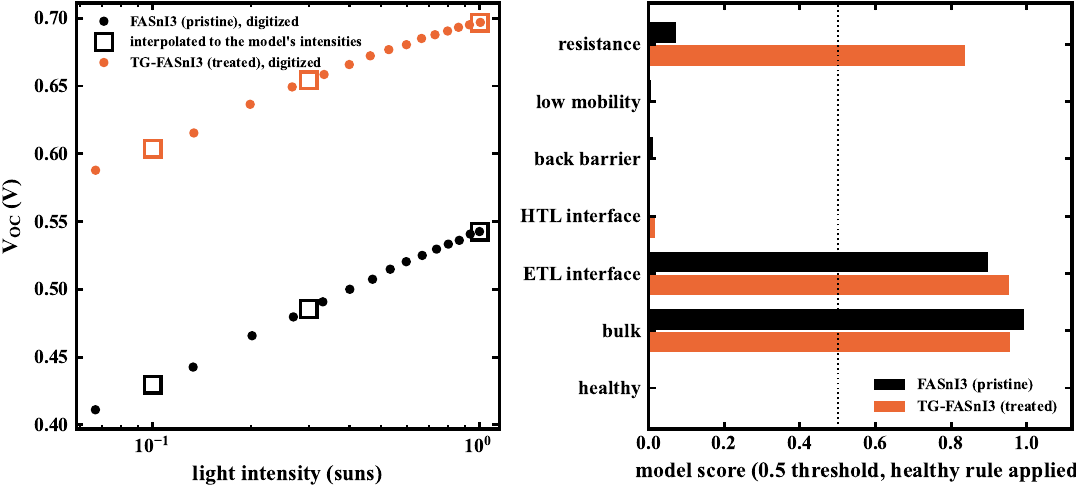}
\caption{Application to two published FASnI\textsubscript{3} cells \citep{liu2020}. Left: digitized
$\Voc(\Phi)$ points and their interpolation onto the three intensities used for training. Right:
resulting class scores. Bulk recombination scores highest for both cells, but an interface term
scores almost as highly, so the diagnosis does not identify a single mechanism.}
\label{fig:measured}
\end{figure}

Applied to the two measured cells, the model assigns a high score to bulk recombination (0.99
and 0.96), consistent with the attribution of the losses to defect density by the original
authors. It also assigns high scores to the ETL interface (0.90 and 0.95) and, for the treated
cell, to series and shunt resistance (0.84). The model therefore does not identify a single
mechanism for these devices, but indicates bulk recombination together with at least one
interface contribution (Fig.~\ref{fig:measured}). These values are model scores rather than
calibrated probabilities of an experimental loss mechanism. An interface contribution is
nevertheless plausible for this material family: \citet{kim2017} measured $\Voc(T)$ on
CH\textsubscript{3}NH\textsubscript{3}SnI\textsubscript{3} cells and attributed their large
$\Voc$ loss to interface rather than bulk recombination. This finding provides independent
evidence that tin perovskites can be interface-limited, but it does not confirm the present
diagnosis, which concerns different devices and cannot be verified without labels.

Quantitative severity does not transfer to the measured devices. A regressor trained on
simulated bulk-limited devices, which reaches $\Rtwo = 0.96$ and an MAE of 0.07 decades in
simulation, predicts \num{9.5e16} and \SI{2.5e16}{\per\centi\metre\cubed} for the two cells. The
published densities, \num{2.89e16} and \SI{5.41e15}{\per\centi\metre\cubed}, were obtained from a
Mott--Schottky $C$--$V$ fit; they are ionized-acceptor densities and \emph{not} the SRH bulk trap
density that the regressor predicts. Because the two are different physical quantities, the
agreement to within a factor of 3--5 should not be interpreted as a validation.

Several analysis choices were required, namely the source of $\Voc$ (the intensity curve or the
$J$--$V$ table), the assumed band gap of FASnI\textsubscript{3} and the intensity window. Across
these choices, the prediction for the treated cell varies from
\SIrange{8.2e15}{4.4e16}{\per\centi\metre\cubed}, a factor of 5.4. Digitization noise is not the
limiting factor, since resampling the digitized points with \SI{5}{\milli\volt} of scatter
changes the predictions by less than 2\%. The severity prediction is therefore regarded as
validated in simulation only, and same-material calibration is recommended before any
quantitative use.

A second, more demanding test examined whether the \emph{ratio} of trap densities between an
untreated and a treated cell transfers to measured devices. For this test, a model trained on the
original and hard datasets (3,120 devices) was used to predict $\log_{10}(N_{t,A}/N_{t,B})$
directly. Three success criteria were fixed before the experiment, so that the outcome could not
be judged retrospectively: agreement with the published ratio within a factor of 2, a
digitization-perturbation interval narrower than a factor of 3, and agreement between feature
sets within a factor of 2. The predicted ratios exceed the published ones by factors of 4.6 and
17.7, and when the digitized points are perturbed, the predictions vary over intervals of
$5.6\times$ and $55\times$. Neither of the first two criteria is therefore met
(Table~\ref{tab:criteria}).

\begin{table}[tbp]
\centering
\caption{Sensitivity of the predicted bulk trap-density ratio for two untreated/treated pairs.
Pub.: ratio reported by the original authors. Pred.: ratio predicted from their published
$\Voc(\Phi)$ data. Dev.: factor by which the prediction deviates from the published value.
Range: factor spanned by the prediction when the digitized points are perturbed by
\SI{4}{\milli\volt}. TG denotes the templated-growth treatment of \citet{liu2020}, and SP
denotes the combined solvent and surface engineering of \citet{faheem2025}. Both treatments are
reported by their authors to reduce the trap density.}
\label{tab:criteria}
\begin{tabular}{@{}lrrrr@{}}
\toprule
Pair & Pub. & Pred. & Dev. & Range \\
\midrule
FASnI\textsubscript{3} / TG \citep{liu2020}  & 5.34 & 94.6 & $17.7\times$ & $55\times$ \\
FASnI\textsubscript{3} / SP \citep{faheem2025} & 2.38 & 10.8 & $4.6\times$ & $5.6\times$ \\
\bottomrule
\end{tabular}
\end{table}
The third criterion is met, with the feature sets agreeing within a factor of 1.4--1.9, which
identifies the source of the failure: the limitation lies not in the choice of features but in
the sensitivity of a slope \emph{difference} to read-off error. The perturbation interval widens
from a factor of 13.9 to 55 and then to 684 as the assumed digitization error increases from 2 to
4 and then to \SI{8}{\milli\volt}. On simulated pairs, the same model performs well, with
$\Rtwo = 0.91$, a median error factor of 1.7 and a device-level bootstrap interval for the MAE of
0.33--0.42 decades. The failure therefore lies in the transfer to measured devices rather than in
the model itself.

Two further cautions apply to the published values. First, the two ratios are not the same
physical quantity: \citet{liu2020} obtained theirs from a Mott--Schottky $C$--$V$ fit and
\citet{faheem2025} from the trap-filled-limit voltage of a space-charge-limited-current
measurement, and neither quantity is the SRH bulk trap density represented in SCAPS. Second, the
second pair fails an internal consistency check, since the ideality factors extracted here from
its published $\Voc(\Phi)$ figure (0.91 and 0.83) do not reproduce the values of 1.32 and 1.13
reported in the same paper. This pair therefore cannot serve as ground truth and is reported only
for completeness.

A third device, the mixed-cation lead perovskite CsRbFAMA of \citet{lopez2023}, was measured by its authors between 253 and \SI{333}{\kelvin}. Because the authors performed the same $\Voc(T)$ extrapolation and reported their own value, this device provides a check on the
feature extraction itself. Fitting between 253 and \SI{313}{\kelvin}, we obtain
$\Ea = \SI{1.497}{\electronvolt}$, compared with their value of \SI{1.49}{\electronvolt}; the
difference of \SI{7}{\milli\electronvolt} confirms that the digitization and the fit reproduce a
published analysis.

The choice of fitting window also proved important. The authors report that bulk SRH
recombination dominates up to \SI{323}{\kelvin} and that surface recombination dominates above
this temperature. Fitting across this transition degrades the fit, with $\Rtwo$ decreasing from
0.988 to 0.941, and raises the extrapolated $\Ea$ from 1.497 to \SI{1.567}{\electronvolt},
approaching the band gap of \SI{1.609}{\electronvolt}. When the raw digitized points are used
directly, without first interpolating the four measured curves onto a common temperature axis, the
extrapolated value exceeds the band gap. The window that overlaps the simulated training range
(\SIrange{300}{400}{\kelvin}) contains only three usable measured points, too few for a reliable slope,
so the overlap between this measurement and the training domain is too narrow to be exploited.

The diagnosis for this device is not stable enough to be reported as a result: resampling the
digitized points with \SI{5}{\milli\volt} of scatter in $\Voc$ and 0.5 points in FF gives a
back-barrier score of $0.74 \pm 0.32$ and a low-mobility score of $0.31 \pm 0.36$. This device is
accordingly reported as a validation of the feature extraction rather than of the diagnosis.

Two general requirements follow from these results. First, the extrapolation must be performed
over a range in which $\Voc(T)$ is linear, because a curved fit invalidates the extrapolated
value. Second, this requirement applies to the manual reading and not to the learned model: as
shown in Section~\ref{sec:protocol}, curvature is itself the signature of a contact barrier, so
excluding non-linear devices from the model would remove the class that it diagnoses most
reliably.
\subsection{A protocol for applying the method to measured devices}
\label{sec:protocol}

The robustness experiments lead to the following practical requirements.

\begin{itemize}
\item \textbf{Train with realistic measurement noise.} A model trained without noise falls below
  the physics rules at \SI{3}{\milli\volt} of $\Voc$ noise, which is within normal experimental
  scatter.
\item \textbf{Match the measurement range, and extract the features identically for measurement
  and simulation.} The features depend on the measurement range and sampling: a curve measured
  at 14 linearly spaced intensities and a curve sampled at three logarithmically spaced
  intensities do not yield the same ideality factor or curvature, even when the underlying
  physics is identical. Measured data should therefore be interpolated onto the intensities used
  for training before the features are extracted. In this work, changing the analyzed range from
  0.1--1 sun to 0.03--1 sun altered a predicted defect density by two orders of magnitude.
\item \textbf{Check linearity before a manual extrapolation, but do not exclude non-linear
  devices from the model.} For a manual determination of $\Ea$, curvature is disqualifying: the
  mixed-cation device of \citet{lopez2023} changes its dominant recombination mechanism above
  \SI{323}{\kelvin}, and fitting across this transition yields an unphysical activation energy
  above the band gap. The same criterion could be applied as a data-quality filter for the model,
  and such a filter was initially considered, but the data do not support it, because curvature
  in $\Voc(T)$ is itself diagnostic. Contact-barrier devices have a median fit quality of
  $\Rtwo = 0.94$, compared with 0.998 for every other class, and constitute 85\% of the least
  linear fifth of the test set. Removing these devices discards the class that the model
  identifies most reliably (AUC 1.000) and lowers the accuracy on the remaining devices, with the
  macro $F_1$ decreasing from 0.830 to 0.793 under noise.
\item \textbf{Abstain on the basis of model confidence, not fit quality.} Withholding a diagnosis
  for the 20\% of devices with the lowest predicted probability raises the macro $F_1$ on the
  remaining devices from 0.830 to 0.845 under measurement noise. Abstention based on the standard
  error of the ideality factor performs similarly (0.843), whereas abstention based on the
  standard error of $\Ea$ gives no improvement (0.813).
\item \textbf{Treat cross-material application as exploratory.} A model trained only on simulated
  \absorber{} devices assigned its highest score to bulk recombination for both measured
  FASnI\textsubscript{3} cells, in agreement with the attribution of the original authors, but it
  assigned an almost equally high score to an interface term, and the predicted defect densities
  varied by a factor of five across defensible analysis choices. Two devices without verified
  labels cannot establish whether the classification transfers, and same-material calibration is
  required before the severity output is used.
\end{itemize}

\subsection{Limitations}

The simulations are one-dimensional and therefore omit lateral inhomogeneity, grain-boundary transport and ion migration, the last of which is interface-dependent in halide perovskites \citep{levine2016} and cannot be represented in one dimension even in principle. Material parameters are held constant with temperature, as is common in SCAPS studies, so computed activation energies should be compared with measurements in relative rather than absolute terms.
Conventional defaults such as the capture cross-sections (\SI{1e-15}{\centi\metre\squared}) and thermal velocities (\SI{1e7}{\centi\metre\per\second}) were not subjected to a sensitivity
analysis. The class set is also closed: a device limited by an unmodeled mechanism, such as the ion migration and temperature-dependent HTL conductivity reported for the measured lead perovskite device, is still assigned to the nearest class, and the present novelty check, based on feature distributions, does not detect such cases.

The sampling result of Section~\ref{sec:sampling} is reproduced in the supplementary material for four further absorbers with band gaps from \SIrange{1.31}{2.00}{\electronvolt} in the same device stack, and selection by grid score chooses the worst-generalizing model in every case. Generality beyond SCAPS-1D and this architecture is, however, not established. The \SI{2.00}{\electronvolt} parameter set of \citet{cspbi3_cssni3_2026}, well above the band gap usually reported for CsSnI\textsubscript{3}, is used without modification as the wide-gap endpoint, and no material-specific conclusion is drawn from it. Finally, all real-device results rely on data digitized from published figures, and in one case the published ideality factors could not be
reproduced from the plotted data.

\subsection{Outlook}

The natural next step is calibration on devices measured in-house, where the ground truth is controlled and both $\Voc(T)$ and $\Voc(\Phi)$ can be recorded on the same cell. Extending the class set to transport-layer degradation and ion migration would cover mechanisms that were encountered in the measured devices but could not be represented here. Finally, repeating the diagnostic analysis for several absorbers would establish whether a single model can serve a
family of thin-film technologies or whether material-specific training is required.

\section{Conclusions}

This work combined SCAPS-1D device simulation with machine learning for lead-free \absorber{} solar cells in both the forward and the inverse directions. The contribution is methodological rather than a new efficiency record. Replacing Spiro-OMeTAD with CuI raises the simulated efficiency from 19.6\% to 21.6\% at identical absorber quality, almost entirely through the fill factor. A surrogate-guided search then identified a high-performance region of the design space
and showed it to be a plateau. The best SCAPS-verified candidate reaches 22.94\%, matching rather than exceeding the best of the 4,096 grid points, although it lies in a different region of the design space.

Two methodological results extend beyond this device. The first concerns the evaluation of surrogate models. Models trained on factorial grids report accuracies that do not hold between the grid points: in this work, one model reached $\Rtwo = 0.9997$ on held-out grid points but only 0.579 on independent devices, whereas a space-filling design of comparable size achieved $\Rtwo = 0.995$. Model selection within the grid aggravates the problem, because it favors the model that generalizes worst.

The second result concerns the inverse direction. The degradation mechanisms present in a simulated cell can be identified from temperature- and intensity-dependent electrical characteristics, with the classifier reaching a macro $F_1$ of 0.938, compared with 0.687 for calibrated physics rules, and forced-choice accuracies of 99\% and 61\%, respectively, on single-degradation devices. The advantage over the rules is largest for devices with more than one degradation mechanism, but it holds only for combinations of mechanisms that are represented
in the training set. For a pairing excluded from training, the model reports the mechanism with the stronger electrical signature and misses the other, an outcome that a controlled parameter sweep predicts correctly for every pair tested.

Finally, the framework was applied to published measurements of two FASnI\textsubscript{3} devices. The classifier assigned its highest score to bulk recombination, in agreement with the attribution of the original authors, but it also indicated an interface contribution. With two devices and no verified labels, this result constitutes an exploratory cross-material application rather than an experimental validation, and the quantitative severity estimates proved sensitive to analysis choices and require same-material calibration.
\section*{Author Contributions}
N.S.: conceptualization, methodology, software, investigation, formal analysis, visualization, writing -- original draft. H.I.: supervision, methodology, writing -- review and editing.

\section*{Notes}
The authors declare no competing financial interest.

\section*{Supporting Information}
The Supporting Information is available free of charge.
\begin{itemize}
  \item Supporting Information (PDF): sampling-design results for four further absorbers, additional dataset, feature and model details, and supplementary figures and tables.
\end{itemize}

\section*{Acknowledgments}
The authors thank the Department of Electrical and Electronic Engineering, Bangladesh University of Engineering and Technology (BUET), for the institutional and computational support that made this work possible. The authors also gratefully acknowledge Prof.~Marc Burgelman and colleagues at the Department of Electronics and Information Systems (ELIS), Ghent University, for developing SCAPS-1D and making it freely available to the photovoltaic research community.

\section*{Data Availability Statement}
\label{sec:data}

The simulation datasets, the SCAPS definition files and the analysis code are available at [repository DOI to be added on acceptance]. Each quantitative result reported here is written to a named CSV file by the script that produces it, so every number in the text and every figure can be traced to the run that generated it. The complete simulation ledger, listing every batch and its run count, is provided as \texttt{scaps\_run\_ledger.csv} and is regenerated by \texttt{run\_ledger.py}.
\bibliography{refs}

\newpage
\section*{TOC Graphic}
\noindent\includegraphics[width=3.25in,height=1.75in]{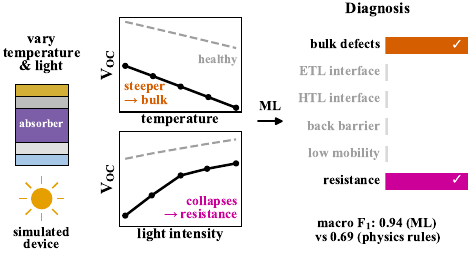}

\end{document}